%% file: sample-sigconf.tex
\documentclass[sigconf]{acmart}

\AtBeginDocument{%
  }

\input{packages}

\definecolor{myblue}{HTML}{0b5394}
\definecolor{myred}{HTML}{c21313} % 980000 c21313 
\definecolor{mygreen}{HTML}{3aa135} % 44a83e 4cbd46 3aa135

\hypersetup{
    colorlinks = true,
    citecolor = red, 
    linkcolor = blue, 
    urlcolor = cyan 
}

\crefformat{equation}{\textcolor{myblue}{\hyperref[#1]{Eq.~#2#1#3}}}
\crefformat{table}{\textcolor{myblue}{\hyperref[#1]{Tab.~#2#1#3}}}
\crefformat{section}{\textcolor{myblue}{Sec.~#2#1#3}}
\crefformat{subsection}{\textcolor{myblue}{Sec.~#2#1#3}}
\crefformat{algorithm}{\textcolor{myblue}{Algorithm~{#2#1#3}}}
\crefformat{figure}{\textcolor{myblue}{\hyperref[#1]{Fig.~#2#1#3}}}

\copyrightyear{2024}
\acmYear{2024}
\setcopyright{acmlicensed}
\acmConference[MM '24]{Proceedings of the 32nd ACM International Conference on Multimedia}{October 28-November 1, 2024}{Melbourne, VIC, Australia}
\acmBooktitle{Proceedings of the 32nd ACM International Conference on Multimedia (MM '24), October 28-November 1, 2024, Melbourne, VIC, Australia}
\acmISBN{979-8-4007-0686-8/24/10}
\acmDOI{10.1145/3664647.3681592}

\begin{document}

\title{Evolution-aware VAriance (EVA) Coreset Selection for Medical Image Classification}


% \settopmatter{authorsperrow=4}

\author{Yuxin Hong}
\orcid{0009-0002-3340-7731}
\affiliation{
  \institution{South-Central Minzu University}
  \department{College of Computer Science}
  \department{Key Laboratory of Cyber-Physical Fusion Intelligent Computing, State Ethnic Affairs Commission}
  \city{Wuhan}
  \country{China}
}
\email{yuxinh@scuec.edu.cn}

\author{Xiao Zhang}
\orcid{0000-0002-8020-6142}
\authornote{Corresponding author}
\affiliation{
  \institution{South-Central Minzu University}
  \department{College of Computer Science}
  \department{Key Laboratory of Cyber-Physical Fusion Intelligent Computing, State Ethnic Affairs Commission}
  \city{Wuhan}
  \country{China}
}
\email{xiao.zhang@my.cityu.edu.hk}

\author{Xin Zhang}
\orcid{0009-0003-3706-9044}
\affiliation{
  % \institution{XiDian University, Xi'an, China}
  % \country{}
  \institution{XiDian University}
  \city{Xi’an}
  \country{China}
}
\affiliation{
  \institution{Agency for Science, Technology and Research (A*STAR)}
  % \department{Centre for Frontier AI Research (CFAR), Institute of High Performance Computing (IHPC)}
  \department{Centre for Frontier AI Research (CFAR)}
  \department{Institute of High Performance Computing (IHPC)}
  \country{Singapore}
}
% \email{xinzhang_xd@163.com}
\email{xinzhang01@stu.xidian.edu.cn}

\author{Joey Tianyi Zhou}
\orcid{0000-0002-4675-7055}
\affiliation{
  \institution{Agency for Science, Technology and Research (A*STAR)}
  % \department{Centre for Frontier AI Research (CFAR), Institute of High Performance Computing (IHPC)}
  \department{Centre for Frontier AI Research (CFAR)}
  \department{Institute of High Performance Computing (IHPC)}
  \country{Singapore}
}
\email{Joey_Zhou@cfar.astar.edu.sg}

\renewcommand{\shortauthors}{Yuxin Hong, Xiao Zhang, Xin Zhang, and Joey Tianyi Zhou}

\newcommand{\Methods}{Evolution-aware VAriance (EVA)}
\newcommand{\Method}{Evolution-aware VAriance}
\newcommand{\method}{EVA }

\input{paragraphs/abstract}






\maketitle

\newcommand{\fix}{\marginpar{FIX}}
\newcommand{\new}{\marginpar{NEW}}

\newcommand{\format}[2]{\begin{tabular}{@{}c@{}}$#1$\\[-3pt]{\color{gray}$\scriptscriptstyle  \pm  #2$}\end{tabular}}
\newcommand{\bformat}[2]{\color{black} \begin{tabular}{@{}c@{}}$\mathbf{#1}$\\[-3pt]$\scriptscriptstyle \mathbf{\pm #2}$\end{tabular}}

% start my content
\input{paragraphs/introduction}
\input{paragraphs/preliminary}
\input{paragraphs/method}
\input{paragraphs/experiment}
\input{paragraphs/related_work}
\input{paragraphs/limitation_future}
\input{paragraphs/conclusion}

\begin{acks}
This work was supported in part by the Fund for Academic Innovation Teams and Research Platform of South-Central Minzu University (Grant Number: XTZ24003, PTZ24001), Knowledge Innovation Program of Wuhan-Basic Research (Project No.: 2023010201010151), and the Research Start-up Funds of South-Central Minzu University under grant YZZ18006, and the Spring Sunshine Program of Ministry of Education of the People's Republic of China under grant HZKY20220331.

Joey Tianyi Zhou is the SERC Central Research Fund (Use-inspired Basic Research). 
\end{acks}

\bibliographystyle{ACM-Reference-Format}
\balance
\bibliography{sample-sigconf}

\clearpage

\title{Supplementary Material}

\appendix

\setcounter{table}{0}
\setcounter{figure}{0}
\renewcommand{\thetable}{A\arabic{table}}
\renewcommand{\thefigure}{A\arabic{figure}}

\input{supplementary/baseline}
\input{supplementary/high}
\input{supplementary/small}
\input{supplementary/cifar}
\input{supplementary/cross}
\input{supplementary/entire}
\input{supplementary/parameter}
\input{supplementary/time}

\end{document}

%% file: packages.tex
\usepackage{url}
\usepackage{bm}
\usepackage{cleveref}

\usepackage{algorithm}
\usepackage{algorithmic}
\usepackage{setspace}
\usepackage{booktabs}

\usepackage{colortbl}
\usepackage{multirow} 

\usepackage{graphicx}  %插入图片的宏包
\usepackage{float}     %设置图片浮动位置的宏包
\usepackage{caption}
\usepackage{subfig}     % 子图包，不要与{subfigure}混用，{subfig}较新
\usepackage{overpic}     % 叭啦叭啦，与图片排版相关

\usepackage{pdfpages}
\usepackage{enumitem}

\usepackage{appendix}
\usepackage{multicol}
\usepackage{balance}

%% file: paragraphs/abstract.tex
\begin{abstract}
  % The high dimensionality and large volumes of medical imaging data necessitate substantial storage, transmission, and computational resources, posing challenges in resource-limited environments such as remote medical facilities and mobile devices. Coreset selection is a promising technique to achieve effective dataset compression. 
  In the medical field, managing high-dimensional massive medical imaging data and performing reliable medical analysis from it is a critical challenge, especially in resource-limited environments such as remote medical facilities and mobile devices. This necessitates effective dataset compression techniques to reduce storage, transmission, and computational cost.
  However, existing coreset selection methods are primarily designed for natural image datasets, and exhibit doubtful effectiveness when applied to medical image datasets due to challenges such as intra-class variation and inter-class similarity. 
  In this paper, we propose a novel coreset selection strategy termed as \textit{\textbf{Evolution-aware VAriance (EVA)}}, which captures the evolutionary process of model training through a dual-window approach and reflects the fluctuation of sample importance more precisely through variance measurement. 
  Extensive experiments on medical image datasets demonstrate the effectiveness of our strategy over previous SOTA methods, especially at high compression rates. EVA achieves 98.27\% accuracy with only 10\% training data, compared to 97.20\% for the full training set. None of the compared baseline methods can exceed Random at 5\% selection rate, while EVA outperforms Random by 5.61\%, showcasing its potential for efficient medical image analysis.
\end{abstract}

%% file: paragraphs/introduction.tex
\section{Introduction}\label{sec:introduction}
In the medical field, data collection and processing are essential for delivering accurate and reliable diagnoses and treatment plans. 
Medical imaging data, typically characterized by high dimensionality and large volumes, necessitates substantial resources for storage and transmission. 
Moreover, training deep learning models on large-scale medical image datasets requires extensive computational resources and time. 
This presents challenges in resource-limited settings, such as remote medical facilities where effective medical image analysis is crucial, or on mobile devices where real-time monitoring and analysis are needed. 
Therefore, efficient data compression and processing techniques become imperative. % necessary
In this context, coreset selection, or dataset pruning, emerges as a promising approach to mitigate these challenges.
% which emerged to reduce data storage and neural network model training cost, serves as a promising solution to address the above obstacles.
Coreset selection condenses a given large-scale dataset into a significantly smaller subset, known as the coreset. 
The coreset is expected to preserving the essential knowledge of the original full dataset such that the former yields a similar performance as the latter. %Thus the model trained on the coreset can achieve performance comparable to that trained on the original dataset.
% \captionsetup{font=small}
\begin{figure}[]
	\centering
        \setlength{\abovecaptionskip}{0.1cm}
	\includegraphics[width=.99\linewidth]{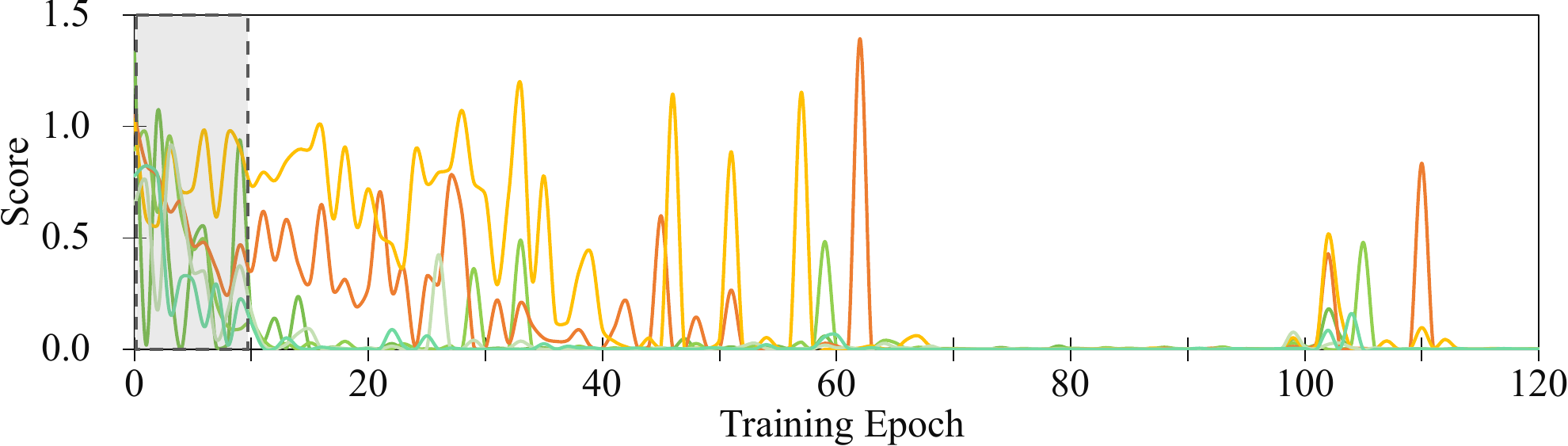}\\
	\includegraphics[width=.99\columnwidth]{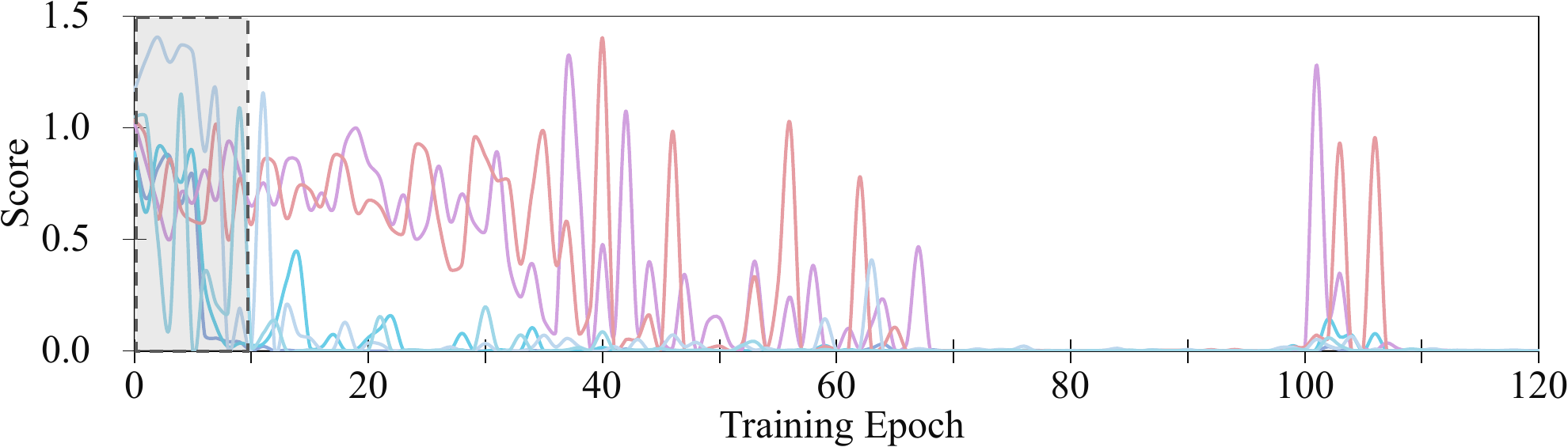}
        \caption{Existing single-timeframe/window snapshots methods fail to capture sample importance fluctuations across epochs. Different samples are denoted in different colors. Here, we measure importance score using the error vector score, a snapshot-based criterion defined in \textcolor{myred}{\cite{paul2021deep}}, considering only the first 10 epochs as indicated by the dashed box. These scores are obtained by training ResNet-18 on OrganAMNIST.}
    \label{fig:fluc}
    \vspace{-0.6cm}
\end{figure}

Numerous coreset selection works \textcolor{myred}{\cite{huang2023efficient, loshchilov2015online, marion2023less, park2024robust, paul2022lottery, xia2022moderate, yang2023towards}} have explored various criteria for identifying important data samples, including geometry distance \textcolor{myred}{\cite{sener2017active, welling2009herding}}, uncertainty \textcolor{myred}{\cite{coleman2019selection}}, loss \textcolor{myred}{\cite{toneva2018empirical, paul2021deep}}, decision boundary \textcolor{myred}{\cite{ducoffe2018adversarial, margatina2021active}}, and gradient matching \textcolor{myred}{\cite{mirzasoleiman2020coresets}}. 
However, most of these methods have been validated mainly on natural image datasets, such as CIFAR-10, CIFAR-100 \textcolor{myred}{\cite{krizhevsky2009learning}}, and not extensively on medical datasets.
% lack comprehensive validation on medical datasets. 
% The effectiveness of existing methods on medical image datasets is doubtful, due to the distinct characteristics and distributions between natural and medical image datasets. 
The applicability of those methods for medical image datasets are under exploration, given the unique characteristics of medical images.
Compared to natural image datasets, the intra-class variation and inter-class similarity of medical image datasets \textcolor{myred}{\cite{song2015large}} pose specific challenges to coreset selection.
%\cref{fig:medical} illustrates the challenges of intra-class variation and inter-class similarity in medical images. 
On the one hand, in medical imaging, samples within the same category can exhibit significant differences, making it difficult to capture consistent features for each class. 
This variation largely comes from the diversity in disease manifestation across patients and discrepancies in imaging conditions.
On the other hand, the challenge of inter-class similarity arises when images representing different diseases exhibit similar visual characteristics. % For example, certain benign and malignant tumors may appear similar in imaging studies, complicating the task of accurate classification. 
\cref{fig:medical} provides a more straightforward demonstration of this characteristic.
These factors contribute to the complexity of medical image analysis and underscore the need for sophisticated coreset selection methods that can effectively address these challenges.

%由于医疗图像数据集和自然图像数据集具有不一样的特点，两种图像上的分类任务要求和策略也不一样
% 保持波动性特征：medical image presents fluctuating characteristics, over time and specific events change (\textcolor{myred}{\cite{cai2020review}). 由于医学影像的波动性，核心集选择算法需要能够捕捉并保留不同阶段和条件下的关键特征。在高压缩率下，选择一个代表性的小子集来维持原始数据集的动态特性是一个挑战。

% 1. evolution: dual-window
% coreset要用更少的样本模拟原始数据集的效果，就要模拟在original dataset上的训练过程(need to simulate model training dynamic on original dataset). 网络学习的过程是一个从易到难的进化演变过程，一开始学一些general easy features, 然后学特异性的hard feature. (\textcolor{myred}{\cite{yosinski2014transferable} 网络浅层学习通用特征，而深层学习特定于任务的特征). For example, kidney image: model一开始学习肾脏的形状，之后再学不同肾脏的细节特征。此外，样本的重要性是随训练阶段变化的（different sample在不同训练阶段发挥作用）\textcolor{myred}{\cite{chang2017active}.
To enable the coreset to effectively approximate the model's performance on the full training set with fewer samples, it is essential to consider the training process on the original dataset. 
This necessitates that the selection methods should effectively capture the varying importance of samples  at different training stages. 
Yosinski et al. \textcolor{myred}{\cite{yosinski2014transferable}} highlighted that shallow layers of the network learn general features, while deeper layers learn task-specific features. 
Han et al. \textcolor{myred}{\cite{han2018co}} observed that deep models tend to memorize easy instances initially and adapt to harder instances as training progresses. 
These studies confirm the evolutionary nature of deep learning from simpler to more complex stages. %Specifically, in the initial stages of training, the model typically learns some general and relatively simple features. As training progresses, the model gradually learns more complex and specific features. For example, in kidney images, the model might first learn to recognize the general shape of the kidney, followed by distinguishing detailed features of different kidneys.
Given these observations\textcolor{myred}{\cite{yosinski2014transferable,han2018co}}, we posit that in the domain of medical imaging, the training process of deep learning models exhibits similar characteristics. 
For instance, in kidney images, the model initially learns the general kidney shape and gradually distinguishes more detailed features of different kidneys.
Moreover, as shown in \cref{fig:fluc}, the significance of samples in enhancing the model performance varies across different training stages \textcolor{myred}{\cite{toneva2018empirical, chang2017active, zhang2023spanning, katharopoulos2018not}}. Specifically, certain samples may be crucial for the model's initial learning phase, while others gain importance in the later stages of training.

Most of the existing methods evaluate sample importance using a snapshot of training progress. For example, % Paul et al. \textcolor{myred}{\cite{paul2021deep} use the error vector score generated by a few epochs in early training; 
Xia et al. \textcolor{myred}{\cite{xia2022moderate}} calculate the distribution distances of features at the end of training. 
Zhang et al. \textcolor{myred}{\cite{zhang2023spanning}} have proved that the importance scores of samples varies with epochs during training, resulting in significant variations in the constructed coresets at different snapshots. 
Therefore, methods reliant on single-timeframe snapshots might be inadequate for capturing the comprehensive evolution of model training, overlooking the dynamic characteristics of learning process. %single-timeframe/window v.s. dual window
% Therefore, snapshot-based methods fail to capture the evolution of model training. 
%This limitation hinder the effectiveness of these snapshot-based methods in constructing a generalized coreset required in high compression rate scenarios. 
% Another challenge arises from the over-reliance on importance scores from individual snapshots, which can lead to overfitting in coreset construction. In scenarios with varying selection rates, the lack of generalization of these snapshot-based approaches becomes apparent. The accuracy gap between coresets constructed at different snapshots widens as the selection rate decreases (see \cref{fig:line_phase}), highlighting the difficulty of generalizing across diverse scenarios.

Expanding the scope of the considered training dynamics is a straightforward approach to address this limitation. Previous studies have attempted %to incorporate training dynamics 
this
using various methods. For example, Pleiss et al. \textcolor{myred}{\cite{pleiss2020identifying}} measures the probability gap between the target class and the second-largest class in each epoch; 
Paul et al. \textcolor{myred}{\cite{paul2021deep}} utilize the expected value of error vector scores generated by 
% a few epochs in early training (the first 10 epochs).
the first 10 epochs.
While this approach partially expands the scope of the considered training dynamics, it overlooks the potential effectiveness of later stages of training, and more importantly, it focuses on samples with high expected error values, indicating that these samples are consistently mispredicted over many training iterations. Such samples may just be too difficult/noisy and may degrade the model performance \textcolor{myred}{\cite{chang2017active}}.
Toneva et al. \textcolor{myred}{\cite{toneva2018empirical}} count the number of forgetting events during training, which occur when samples, previously classified correctly, are subsequently predicted incorrectly. However, this counting approach only provides the discrete probability of an event, lacking granularity to reflect the variations of sample contributions throughout the training process. 

To address these limitations, in this paper, we propose a novel sample importance scoring strategy called \textbf{\Methods}, aiming at achieving reasonable and effective compression of medical image datasets. % aiming at overcoming the limitations of existing coreset selection methods for medical image datasets. % 首先，为了避免只考虑snapshot或者一段训练过程带来的偏差，我们consider training dynamics of different stage. epochs in dual-window, representing early and late stage of training process. 更好地捕捉模型学习的evolution过程。其次，inside each window，为了用更precise的方法反映样本重要性在模型学习过程中的波动，我们计算样本误差的方差。
Firstly, to mitigate the biases from focusing solely on a snapshot or single segment of the training process, we introduce a dual-window approach that considers training dynamics at different stages. This strategy provides a more holistic understanding of the model's learning evolution, enabling nuanced assessment of sample importance as the model evolves from learning general to specific features. % To capture the dynamic nature of deep learning and avoid biases associated with snapshot-based methods, we introduce a dual-window approach that considers training dynamics at both early and late stages. This strategy enables a comprehensive view of the model's learning evolution, allowing us to accurately assess the changing importance of samples as the model progresses from learning general features to more specific ones.
Secondly, within each window, to reflect the fluctuation of sample importance during the model training process in a more precise way, we propose to measure the variance of samples' error vector. 
The combination of these two strategies provides a more refined and accurate evaluation of sample importance, enabling a more effective coreset selection that aligns with the dynamic nature of neural network training. 
This approach is particularly beneficial in high compression scenarios for medical image datasets, where maintaining accuracy and reliability is challenging but crucial.

In a nutshell, our contributions can be summarized as follows.
\setlist[itemize]{leftmargin=3mm}
\begin{itemize}
\setlength{\itemsep}{-1pt}
\setlength{\parsep}{-1pt}
\setlength{\parskip}{-1pt}
    \item 
    % \textbf{any experimental proof or theoretical proof?}
    We identify the limitations of existing coreset selection methods in capturing the evolutionary nature of model training and the fluctuations in sample importance within medical image datasets.
\end{itemize}
\begin{itemize}
    \item 
    % \textbf{dual-window approach, variance measurement. You would consider using two sentences to describe them, respectively.}
    We thereby propose a novel coreset selection strategy called \textbf{\Methods}, which features two key components. The first is a dual-window approach that captures the training dynamics by considering distinct stages of the learning process. The second is the employment of variance measurement on samples' error vectors, offering a granular and more precise evaluation of each sample's contribution to the model training.     
    % which incorporates a dual-window approach to consider training dynamics at different stages and employs variance measurement of samples' error vectors for a more precise evaluation of sample importance.
\end{itemize}
\begin{itemize}
    \item 
    % \textbf{only one dataset?}
    Extensive evaluations on the OrganAMNIST and OrganSMNIST datasets demonstrate that our \method strategy outperforms SOTA methods at challenging low selection rates while achieving comparable accuracy at high selection rates, showcasing its potential for efficient medical image analysis.
\end{itemize}

%% file: paragraphs/preliminary.tex
\section{Preliminaries}\label{sec:preliminaries}

% \subsection{Preliminaries}\label{sec:preliminaries}

In this paper, vectors and matrices are denoted by bold-faced letters. Given a large-scale dataset, we denote the full training set contains N samples as $\mathbb{T}=\left\{\left(\bm{x}_n, \bm{y}_n\right)\right\}_{n=1}^N$, where $\bm{x}_{n} \in \mathbb{R}^{D}$ represents the input feature vector and the corresponding ground-truth label is $\bm{y}_{n} \in \mathbb{R}^{1 \times C}$, $C$ is the number of classes. All samples are drawn i.i.d. from a underlying distribution $\mathcal{P}$.
% $\bm{y}_{n} \in \mathcal{Y}=\left\{ 0,1, ..., C-1 \right\}$
We define the neural network as $f_{\bm{\theta}}$, parameterized by the weight vector $\bm{\theta}$. The model output $f_{\bm{\theta}}( \bm{x}_{n} ) \in \mathbb{R}^{1 \times C}$ represents the predicted probabilities of each class.
Coreset selection aims to construct a subset (or coreset) $\mathbb{S}=\left\{\left(\bm{x}_m, \bm{y}_m\right)\right\}_{m=1}^M$ ($\mathbb{S} \subset \mathbb{T}$) that captures the essential characteristics of the full dataset, enabling model $f_{\bm{\theta^{\mathbb{S}}}}$ trained on $\mathbb{S}$ to achieve comparable or even superior performance compared to model $f_{\bm{\theta^{\mathbb{T}}}}$ trained on the entire training set $\mathbb{T}$. The data selection rate $\alpha$ in constructing the coreset is then $\frac{M}{N}$. Under these definitions, following previous work \textcolor{myred}{\cite{sener2017active}}, we formulate the objective of coreset selection 
% as the optimization problem: % selection rate: $\frac{| \mathbb{S} |}{| \mathbb{T} |}$
% \begin{equation}\label{eq:optimization}
%     \underset{\mathbb{S} \subset \mathbb{T}:\ \frac{| \mathbb{S} |}{| \mathbb{T} |}=\alpha}{\text{min}}
%     \ \mathbb{E}_{( \bm{x}, \bm{y} )\sim \mathcal{P}}
%     \left[ \ell ( \bm{x}, \bm{y}; f_{\bm{\theta^{\mathbb{S}}}} )  \right],
% \end{equation}
% where $\ell( \cdot  )$ is the loss function, e.g. cross-entropy loss and mean squared error.
as,
\begin{equation}\label{eq:optimization}
\underset{\substack{( \bm{x}, \bm{y} )\sim \mathcal{P} \\ \bm{\theta}_{0}\sim \mathcal{G}}}{\mathbb{E}} \left[ \ell( \bm{x}, \bm{y}; f_{\bm{\theta}_{0}}^{\mathbb{S}} ) \right] 
\simeq 
\underset{\substack{( \bm{x}, \bm{y} )\sim \mathcal{P} \\ \bm{\theta}_{0}\sim \mathcal{G}}}{\mathbb{E}} \left[ \ell( \bm{x}, \bm{y}; f_{\bm{\theta}_{0}}^{\mathbb{T}} ) \right]
\end{equation}
where $f_{\bm{\theta}_{0}}^{\mathbb{S}}$ and $f_{\bm{\theta}_{0}}^{\mathbb{T}}$ represent the neural networks trained on $\mathbb{S}$ and $\mathbb{T}$ with weight $\bm{\theta}_{0}$ initialized from distribution $\mathcal{G}$.

% \subsection{Limitations of Conventional DP methods}\label{sec:limitations}
% Conventional DP methods lack effectiveness on transfer learning / generalization.
% figure (EL2N, ...)

%% file: paragraphs/method.tex
\section{Methodology}
% Story: a novel metric to select samples.
% Variance: Samples with large variance (fluctuate) in the contribution of different epochs to the decline of the training loss.
% 2 Windows: use epochs corresponding to different phases to calculate our score.
% Design (Overall Framework figure). Formulation (explanation). Pseudo-code (Algorithm 1).
% Our method: Evolutionary Importance Sampling

% As discussed in the previous section, while covering a broader range of training dynamics provides a more comprehensive evaluation of individual samples, the conventional averaging technique used in existing methods poses challenges in constructing a coreset with robust generalization. In this section, we embark on introducing some essential preliminaries underlying coreset selection ( \cref{sec:preliminaries}). Subsequently, in  \cref{sec:EVA}, we present the proposed \Methods selection strategy, which aims to strike a delicate balance between capturing training dynamics and identifying well-generalized samples. Additionally, we elucidate the rationale behind the design of two distinct windows in  \cref{sec:2window}. Lastly, we present a detailed exposition of the procedure of \method scoring strategy (see Algorithm \cref{alg:alg}).

% \vspace{-0.2cm} % after Sec.3 title
\begin{figure*}%[htbp]
    \centering
    \setlength{\abovecaptionskip}{0.cm}
    \includegraphics[width=0.88\linewidth]{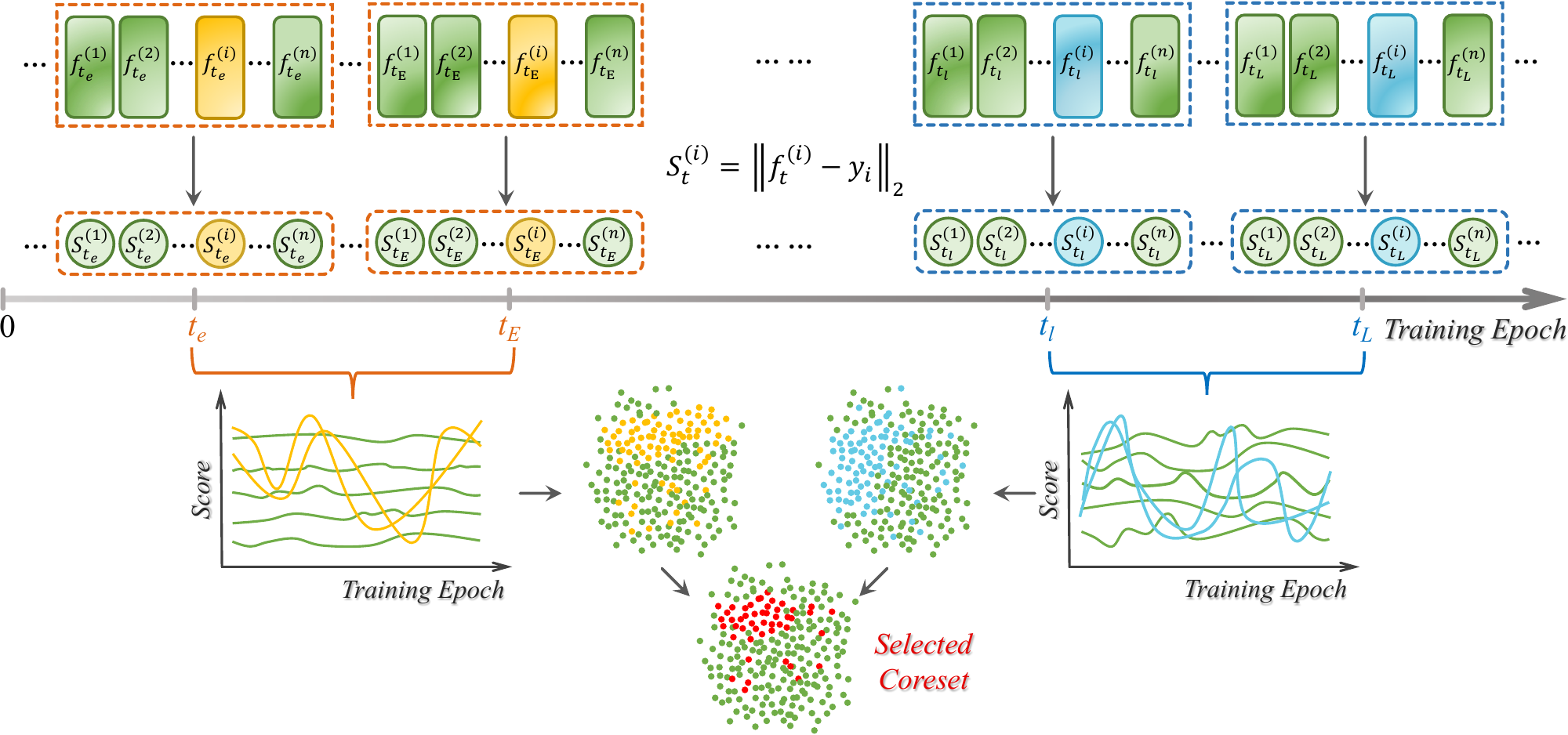}
    % \caption{The Pipeline of our proposed \method.}
    \captionof{figure}{The pipeline of our proposed \method. First, we record individual predicted probabilities ${f}_{t}^{(i)}=f_{\bm{\theta_t}}( \bm{x}_{i} )$ of samples during training. Then, we measure a score $\mathrm{\mathcal{S}}_{t}^{\left( i \right)}$ for each sample, i.e. the L2 norm of error vector. Next, the variance of scores within a window of epochs are calculated to reflect the fluctuation of each sample's contribution. Samples that fluctuate the most are considered important in this stage. Finally, we identify samples that exhibit high importance in dual-window.}
    \label{fig:pipeline}
    \vspace{-0.4cm}  % after this figure
\end{figure*}%, height=0.42\linewidth

To construct a coreset that satisfies \cref{eq:optimization}, the error/loss-based approaches propose to measure the contribution of each sample by considering factors such as the loss, gradient, or its influence on other samples’ prediction during model training \textcolor{myred}{\cite{guo2022deepcore}}. In this context, samples that contribute more to the error or loss are considered more important and are thus selected as part of the coreset. % the reduction in training loss serves as the primary criterion for sample assessment. 

In this section, we delve into the specifics of our proposed \Methods \ strategy, which comprises two key components.
Firstly, we describe how EVA reflects the epoch-level fluctuation by calculating the variance of error-based scores in \cref{subsec:variance}. 
Following that, we elaborate on how EVA captures the training evolution through a dual-window approach in \cref{subsec:2window}.

\subsection{Reflecting Epoch-Level Fluctuation via Variance}\label{subsec:variance}
    % We calculate the L2 norm of the error vector, representing the discrepancy between model predictions and ground-truth labels, at each epoch during training. 
    To approximate the individual contribution of each sample to the reduction in model loss, we initially calculate the variance of error-based scores over a segment of epochs. This process can be further divided into the following steps.
    
    \vspace{1em}
    \textbf{Step 1. Single Epoch Scoring.}
    In this step, we concentrate on calculating the error score for each sample at a specific epoch across multiple independent runs. Specifically, for each sample $( \bm{x}_{i}, \bm{y}_{i} )$ in the training set, we first consider a single epoch $t$ and compute the total mean square error (MSE) across all categories using the equation below: 
    \begin{equation}\label{eq:mse}
        \mathrm{MSE}_{t}^{(i)} = \sum_{j=1}^{C}( \hat{y}_{ij} - y_{ij} )^{2},  
    \end{equation}
    where $\bm{\hat{y}}_{i} = f_{\bm{\theta}}( \bm{x}_{i} )$, therefore $\hat{y}_{ij}$ denotes the model output of the $i$-th sample for the $j$-th category, and $y_{ij}$ is the one-hot encoding of the ground-truth label for the $i$-th sample in the $j$-th category. Then, we take the square root of the total MSE for each sample.
    
    Thus, for each sample $( \bm{x}_{i}, \bm{y}_{i} )$ at epoch $t$, % \textit{Step 1} can be represented as the L2 norm of the error vector:
    we have the L2 norm of the error vector, representing the discrepancy between model predictions and ground-truth labels:
    \begin{equation}\label{eq:score}
        % \mathrm{\mathcal{S}}_{t}^{\left( i \right)} = \left\| f_{\bm{\theta}}\left( \bm{x}_{i} \right) - \bm{y}_{i} \right\|_{2},
        \mathrm{\mathcal{S}}_{t}^{\left( i \right)}=\left\| f_{\bm{\theta}} \left( \bm{x}_{i} \right)-\bm{y}_{i} \right\|_{2},
    \end{equation}
    
    This process yields a sequence of error scores, providing insights into the prediction performance of the model across different training iterations.
    
    \vspace{1em}
    \textbf{Step 2. Variance Across Multiple Epochs.}
    Having obtained the error scores for each sample at individual epochs, in this step, we proceed to assess the variability of scores across multiple epochs by calculating the variance over a segment of epochs. Specifically, for each sample $(\bm{x}_{i}, \bm{y}_{i})$, we analyze the training dynamics over a span of $K$ epochs, from $t$ to $t+K-1$. The error-based scores for this period are represented as     
    $\left\{  {\mathcal{S}_{t}^{(i)}, \mathcal{S}_{t+1}^{(i)}, ..., \mathcal{S}_{t+K-1}^{(i)}} \right\}$.
    %  ${\left\{ \mathrm{\mathcal{S}}_{k}^{(i)} \right\}}_{k=t}^{t+K-1}$
    % The variance of these scores within the $K$-epoch window can be computed as follows:
    We then compute the variance of these scores within the $K$-epoch window using the following equation:
    \begin{equation}\label{eq:variance}
        {\mathcal{V}}_{t}^{(i)} = \frac{1}{K} \sum_{k=t}^{t+K-1} \left( \mathcal{S}_{k}^{(i)} - \mathcal{E}_{t}^{(i)} \right)^2,
    \end{equation} 
    where ${\mathcal{E}}_{t}^{(i)} = \frac{1}{K} \sum \mathcal{S}_{k}^{(i)}$ denotes the mean value within the $K$-epoch window.
    % $\bar{\mathcal{S}_{t}^{(i)} = \frac{1}{K} \sum_{k=t}^{t+K-1} \mathcal{S}_{k}^{(i)}$
    This calculation provides insight into the consistency or variability of the error-based scores for each sample over a specified segment of training epochs, enabling a more precise understanding of the subtle fluctuations in a sample's impact on model performance over time.
    % Note that \cref{eq:variance} can also be expressed in the form of expectation:
    % \begin{equation}\label{eq:var_exp}
    %     % {\mathbb{V}}_{e}^{(i)} = \mathbb{E}[{S_{k}^{(i)}}^2] - (\mathbb{E}[{S}_{k}^{(i)}])^2, \\
    %     % {\mathcal{V}}_{e}^{(i)} = \mathbb{V}[S_{k}^{(i)}] =\mathbb{E}[{S_{k}^{(i)}}^2] - (\mathbb{E}[{S}_{k}^{(i)}])^2, k \in (t_e, t_E), \\
    %     {\mathcal{V}}_{t}^{(i)} = \mathbb{V}[\mathcal{S}_{k}^{(i)}] =\mathbb{E}[{\mathcal{S}_{k}^{(i)}}^2] - \left( \mathbb{E}[\mathcal{S}_{k}^{(i)}] \right)^2,
    % \end{equation}

% \subsection{Capturing Training Dynamics with Dual Window}\label{subsec:2window}
\subsection{Capturing Training Evolution with Dual-Window}\label{subsec:2window}
    % According to \cref{eq:add}, calculating the variance requires two windows of epochs, representing the early and late stages of training, respectively. This design is motivated by two main considerations: 
    As mentioned in \cref{sec:introduction}, snapshot-based methodologies often fall short in capturing the comprehensive evolution of model training, 
    %selecting samples with robust generalization performance, 
    thus warranting an expansion in the scope of considered training dynamics. 
    One approach to broaden the scope is to sample some epochs during the training dynamics. However, random or probabilistic sampling of epochs %\cite{tan2024data}
    may not effectively capture the dynamic changes in sample importance throughout the entire training process. 
    Another method is to consider epochs within a certain window, as we did in \cref{eq:variance}. Nevertheless, this approach carries the risk of excessive bias towards specific training phases.
    
    Therefore, we introduce a dual-window approach to capture the evolution of the training process more comprehensively.
    The first window focuses on the early stages of training, during which the model primarily learns general features. Samples that significantly impact the overall model performance are likely to exhibit high importance during this stage. The second window targets the later stages of training, where the model gradually learns more specific task-related features. The importance of samples that have a significant impact on the overall model performance may increase or decrease during this stage. 
    By integrating information from dual windows, we aim to identify samples that exhibit high importance in both early and late stages. This implies that these samples contain both general features and specific task-related features.
    Additionally, the continuous sequence of epochs provides more temporal information, allowing for a more comprehensive assessment of sample importance throughout the entire training process.
    % At the same time, given that the purpose of coreset selection is to curtail storage requirements and training expenses,     computing the entire training dynamics would incur prohibitive computational and memory costs.
    % Therefore, we set the goal to reflect as accurately as possible the contribution of the samples throughout the training process by observing some slices of the training dynamics. 
    % We aim to avoid focusing solely on either early or late epochs but instead seek to incorporate both, guiding us to design two windows.
    % This design allows coverage of a broader range of different stages of training dynamics while alleviating computational and storage burdens, thus providing the possibility of selecting well-generalization samples. 
    %Focusing solely on either early or late epochs will lead to bias emphasis on partial samples. Thus, we consider both.
    % Compared to considering only one window of training dynamics, e.g., calculating EL2N scores in the early stages of training, the design of two windows offers several advantages. 
    % By examining both the early and late stages of training, the dual-window approach allows for identifying samples that consistently exhibit high importance throughout the entire training process. This enables the detection of subtle changes in sample importance over time, which may not be captured when focusing solely on one period. 
    % Additionally, the inclusion of both early and late stages helps mitigate the risk of excessive bias towards specific training phases, leading to a more balanced and robust assessment of sample contributions. 
    Overall, the use of two windows provides a more nuanced understanding of training dynamics and sample importance, enhancing the effectiveness of the selection process for constructing a coreset. This effectiveness has been proved in  \cref{sec:ablation}.
    
    To maintain consistency with  \cref{subsec:variance}, in the dual-window scenario, we also consider windows spanning $K$ epochs. We define the total number of training epochs as $T$, the first window ranges from $t_e$ to $t_E=t_e+K-1$, and the second window ranges from $t_l$ to $t_L = t_l + K - 1$. These windows are non-overlapping ($t_E < t_l$). Specifically, for each sample $( \bm{x}_{i}, \bm{y}_{i} )$, we compute the scores within each window of epochs, denoted as ${\left\{ \mathrm{\mathcal{S}}_{k}^{(i)} \right\}}_{k=t_e}^{t_E}$ and ${\left\{ \mathrm{\mathcal{S}}_{k}^{(i)} \right\}}_{k=t_l}^{t_L}$. The variance of these scores in \cref{eq:variance} can be formulated as:
    \begin{equation}\label{eq:variance_2windows}
        \begin{split}
         % \mathcal{V}_{t}^{(i)} = \sum_{t-K+1}^{t}\left\| | \mathcal{S}_{t}^{(i)} | - \overline{| \mathcal{S}_{t}^{(i)} |} \right\|^{2},
        {\mathcal{V}}_{e}^{(i)} = \frac{1}{K} \sum_{k=t_e}^{t_E} \left( \mathcal{S}_{k}^{(i)} - {\mathcal{E}}_{e}^{(i)} \right)^2, \\
        {\mathcal{V}}_{l}^{(i)} = \frac{1}{K} \sum_{k=t_l}^{t_L} \left( \mathcal{S}_{k}^{(i)} - \mathcal{E}_{l}^{(i)} \right)^2,
        \end{split}
    \end{equation}
    % where $\overline{| \mathcal{S}_{t}^{(i)} |}$ is the average of score in a window. 
    Here, ${\mathcal{E}}_{e}^{(i)}$ and ${\mathcal{E}}_{l}^{(i)}$ denote the average score of sample $( \bm{x}_{i}, \bm{y}_{i} )$ in two windows, respectively. % Correspondingly, in the expectation form \cref{eq:var_exp}, for early stage, $\mathbb{E}[\mathcal{S}_{k}^{(i)}]$ is equivalent to ${\mathcal{E}}_{e}^{(i)}$ in \cref{eq:variance_2windows}, and $k \in (t_e, t_E)$. For late stage, $\mathbb{E}[\mathcal{S}_{k}^{(i)}] = \mathcal{E}_{l}^{(i)}$, $k \in (t_l, t_L)$. 
    % In this formation, we can intuitively observe the distinction between EVA scores and EL2N.
    
    Finally, we aggregate the variances from both windows to identify samples that demonstrate high importance in two stages. Thus the \method score of each sample can be represented as:
    \begin{equation}\label{eq:add}
        % {\mathbb{V}}^{(i)} = \lambda \mathcal{V}_{t_{E}}^{(i)}+( 1- \lambda) \mathcal{V}_{t_{L}}^{(i)}
        % {\mathcal{V}}^{(i)} = \lambda {\mathcal{V}}_{e}^{(i)}+( 1- \lambda) {\mathcal{V}}_{l}^{(i)}
        {\mathcal{V}}^{(i)} = {\mathcal{V}}_{e}^{(i)}+{\mathcal{V}}_{l}^{(i)}
    \end{equation}
    % where $\lambda$ is a parameter that controls the weight of the variance in the early and late windows.

    % By computing the variance of error scores within each window, we gain a nuanced understanding of how sample importance fluctuates across different training stages, facilitating the identification of well-generalized samples.

    We then sort samples in the full training set $\mathbb{T}$ by their EVA score ${\mathcal{V}}^{(i)}$. Samples with higher scores are deemed more effective at reducing training loss. Given a selection rate $\alpha$, we select the top-ranked M samples to form the coreset, where $M = \left\lceil \alpha N \right\rceil$. % Given a selection rate $\alpha$, the selected coreset can be denoted as $\mathbb{S}=\{\left(\bm{x}_m, \bm{y}_m\right), {\mathcal{V}}^{(m)}\}_{m=1}^{M}$.
    
    \vspace{2pt}
    \cref{alg:alg1} provides a detailed explanation of the procedure for the EVA scoring strategy.
    \input{paragraphs/algorithm}
    

%% file: paragraphs/algorithm.tex
\begin{algorithm}
\caption{\Methods \ Scoring Strategy}
 \label{alg:alg1}
 \setstretch{1.1}
 \begin{minipage}{0.99\linewidth}
   {\textbf{Inputs:}
   Full training set $\mathbb{T}=\left\{\left(\bm{x}_n, \bm{y}_n\right)\right\}_{n=1}^N$; Selection rate $\alpha$; \\ Network $f_{\bm{\theta}}$ with weight $\bm{\theta}$; Epochs $T$; Iteration $I$ pre epoch; \\
   Early window ($t_e$, $t_E$); Late window ($t_l$, $t_L$). 
   % $\mathbb{T}=\left\{\left(\bm{x}_n, \bm{y}_n\right)\right\}_{n=1}^N$: full training set; 
   % $f_{\bm{\theta}}$: network with weight $\bm{\theta}$; \\
   % $\alpha$: selection rate; %$\eta$: learning rate; 
   % $T$: epochs; $I$: iteration pre epoch; \\ %$K$: window size; 
   % ($t_e$, $t_E$): early window; ($t_l$, $t_L$): late window.  
 \begin{algorithmic}[1]
            \FOR{$t=1$ to $T$}
                \FOR{$i=1$ to $I$, sample a mini-batch $\mathbb{B}_i \subset \mathbb{T}$}
                    \STATE Obtain predicted probabilities $f_{\bm{\theta}_t}(\bm{x}_n)$, $\bm{x}_n \in \mathbb{B}_i$
                    \STATE Calculate $\mathcal{S}_{i}^{(n)}$ by defined \cref{eq:score} for each $\bm{x}_n$ \hfill
                    %// Defined in \cref{eq:score}\\
                    \STATE Update $\mathcal{S}_{t}^{(n)}+=\mathcal{S}_{i}^{(n)}$
                \ENDFOR
                \IF{$t_e\leq t < t_E$}
                    % \vspace{2.5pt}
                    \STATE Calculate ${\mathcal{V}}_{e}^{(n)}$ by defined \cref{eq:variance_2windows} of early window, $\bm{x}_n \in \mathbb{T}$
                \ELSIF{$t_l\leq t < t_L$}
                    \STATE Calculate ${\mathcal{V}}_{l}^{(n)}$ by defined \cref{eq:variance_2windows} of late window, $\bm{x}_n \in \mathbb{T}$\hfill % // Defined in \cref{eq:variance_2windows}
                    % \\ \COMMENT{Defined in \cref{eq:variance_2windows}} % \autoref{window_variance}
                \ELSIF{$t = t_L$}
                    \STATE Update ${\mathcal{V}}^{(n)}$ by defined \cref{eq:add} as the \method score of $\bm{x}_n$\hfill % // Defined in \cref{eq:step2}
                % \COMMENT{Define in \cref{eq:step2}}
                \ENDIF
            \ENDFOR
  \STATE Sort samples by ${\mathcal{V}}^{(n)}$ in descending order, $\bm{x}_n \in \mathbb{T}$\hfill
  % Select M samples with highest ${\mathcal{V}}^{(m)}$
  % Select top-M samples $\mathbb{S} \leftarrow\{\left(\bm{x}_m, \bm{y}_m\right), {\mathcal{V}}^{(m)}\}_{m=1}^{M}$ %, M = \left\lceil \alpha N \right\rceil$
 \end{algorithmic}}
 {\textbf{Output:} Top-M samples as the coreset $\mathbb{S}=\left\{\left(\bm{x}_m, \bm{y}_m\right)\right\}_{m=1}^M$ %; associated sample-wise importance $r=\{{\mathcal{V}}^{(m)}\}_{m=1}^{M}$
 }
 \end{minipage}
 \hspace*{0.02in}
\end{algorithm}
% \vspace{-0.3cm}

%% file: paragraphs/experiment.tex
\section{Experiments}
\label{experiments}
\input{paragraphs/tabs/tab_low}

In this section, we present a comprehensive set of experiments and analyses to showcase the effectiveness of our proposed \Method \ scoring strategy in diverse scenarios. We begin by empirically evaluating our \method method against other baselines (\cref{sec:baseline}). Subsequently, we conduct a series of ablation studies to investigate the effectiveness of the proposed two main components: variance measurement and dual-window strategy (\cref{sec:ablation}). Additionally, we perform cross-architecture experiments to evaluate the robustness of our coresets, assessing their performance when selected on one architecture and tested on others.
    \vspace{-0.2cm}
    \subsection{Experiment Setup}\label{subsec:experiment_setting}
    
    \vspace{5pt}
    \textbf{Datasets.} \ \ MedMNIST is a large-scale collection of medical images comprising 10 datasets, covering multi-modal, diverse data scales (from 100 to 100,000) and classification tasks. 
    % The classification performance of this public large-scale datasets 
    Its classification performance 
    has been validated as effective in \textcolor{myred}{\cite{medmnistv2}}. 
    More details % about MedMNIST are included in
    are in \cref{subsec:medical}.
    In this work, due to the time-consuming training, %the effectiveness of the proposed method is primarily evaluated 
    we primarily evaluate our method 
    on two 2D datasets from MedMNIST: OrganAMNIST and OrganSMNIST \textcolor{myred}{\cite{bilic2023liver, xu2019efficient}}, both derived from 3D %computed tomography (CT) 
    CT images from the Liver Tumor Segmentation Benchmark (LiTS). These datasets are designed for multi-class classification tasks, involving 11 body organs with labels including the bladder, femur (left and right), heart, kidney (left and right), liver, lung (left and right), pancreas, and spleen. 
    OrganAMNIST, previously OrganMNIST-Axial in MedMNIST v1 \textcolor{myred}{\cite{yang2021medmnist}}, consists of 58,830 axial view slices of abdominal CT images, 
    % distributed into 34,561 training, 6,491 validation, and 17,778 testing images.
    with 34,561 for training, 6,491 for validation, and 17,778 for testing. 
    OrganSMNIST, formerly OrganMNIST-Sagittal, includes 107,180 abdominal CT images split into 
    % 13,932 training, 2,452 validation, and 8,827 testing images.
    13,932 for training, 2,452 for validation, and 8,827 for testing.

    \vspace{5pt}
    \textbf{Baselines and Networks.} \ \ 
    We compare our method against six representative baselines, the latter five of which are state-of-the-art (SOTA) methods: 1) \textbf{Random}; 
    2) \textbf{Forgetting score}~\textcolor{myred}{\citep{toneva2018empirical}}; 3) \textbf{Entropy}~\textcolor{myred}{\citep{coleman2019selection}}; 4) \textbf{EL2N}~\textcolor{myred}{\citep{paul2021deep}}; 5) \textbf{Area under the margin AUM)}~\textcolor{myred}{\citep{pleiss2020identifying}}; 6) \textbf{Coverage-Centric Coreset Selection (CCS)}~\textcolor{myred}{\citep{zheng2022coverage}}. Details of these baselines are provided in the Supplementary due to space limitations.
    The effectiveness of these strategies is evaluated based on their ability to select representative samples for coreset construction using various criteria. For all baselines except CCS, coresets are formed by pruning less important examples according to the respective importance metric.

    The effectiveness of our method is primarily evaluated using ResNet-18 \textcolor{myred}{\cite{he2016deep}}. We also conduct cross-architecture generalization experiments with ResNet-50 \textcolor{myred}{\cite{he2016deep}}, MobileNet \textcolor{myred}{\cite{sandler2018mobilenetv2}} and LeNet \textcolor{myred}{\cite{lecun1998gradient}} to validate its robustness across different models. Further details are available in the Supplementary.
    % \cref{subsec:cross-architecture}. LeNet, MLP
    % \input{paragraphs/algorithm}
        
    \vspace{5pt}
    \textbf{Implementation details.}
    To ensure fairness in our comparisons, we adhere to the experimental setup outlined in \textcolor{myred}{\cite{zheng2022coverage}}. Our method is implemented using PyTorch \textcolor{myred}{\cite{paszke2017automatic}} and all models are trained on an NVIDIA 3090 GPU. Unless otherwise noted, we utilize the same network architecture, ResNet-18, for both the coreset and the surrogate network on the full dataset. 
    Consistency in hyperparameters and experimental settings is maintaind before and after coreset selection. The surrogate network is trained for 200 epochs across all datasets. Initially, we train a network on the complete dataset to establish baseline performance. We then calculate importance scores by assessing the variance of each sample's error vector across multiple epochs within a dual-window.
    As for the start and end epoch of each window, we employ a grid search with a 10-step size ($K=10$). This process helps us identify the most effective window combinations, denote as ($t_e$, $t_E$)+($t_l$, $t_L$) for different datasets and selecting rate $\alpha$.
    
    % \vspace{-3pt}
    \subsection{Benchmark Evaluation Results}\label{sec:baseline}
        Our systematic comparison of \method against other baselines, as detailed in \cref{subsec:experiment_setting}, reveals its superior performance on the OrganSMNIST and OrganAMNIST medical datasets, particularly at more challenging selection rates. As shown in \cref{tab:low_SA}, our \Method \ approach consistently achieves top-ranking performance, underscoring its robustness in coreset selection. In addition, on the OrganAMNIST dataset, \method nearly matches the full dataset's performance at a 20\% selection rate and surpasses it at 30\%, highlighting its efficiency in utilizing smaller datasets.
        Notably, at extremely low selection rate of 2\% and 5\% on the OrganSMNIST dataset, \method surpasses the Random baseline by a margin of 2.49\% and 5.61\%, respectively, illustrating its effectiveness even with severely limited data, establishing the method's capability to discern and retain the most influential samples for model training. 

        The baselines, including well-established SOTA methods, do not exhibit the same level of performance at these lower selection rates, often failing to exceed the benchmark set by random selection. This trend highlights the limitations of traditional coreset selection methods when dealing with the complexities of medical datasets.
        
        Here, our experiments focus on low selection rates scenarios, but \method also maintains competitive performance at high selection rates. 
        Moreover, our methodology's effectiveness is not confined to medical imaging datasets alone. Preliminary experiments on widely recognized natural image datasets, such as CIFAR, corroborate that \method stands out by surpassing most SOTA methods. 
        Additionally, we investigate the possibility of overfitting on smaller datasets (DermaMNIST and PneumoniaMNIST). 
        Please refer the Supplementary for detailed results of the additional experiments mentioned above.
        % Detailed results from these additional experiments are documented in the Supplementary due to space constraints.

    \vspace{-3pt}
    \subsection{Ablation Study and Analysis}\label{sec:ablation}
        We delve into ablation studies to dissect the contributions of the variance and dual-window components in our method. By systematically removing each component and evaluating their impact on performance, we elucidate their individual roles in enhancing coreset selection accuracy. 
        In this context, we partition our experiments into four conditions: \textsc{Var-S}, \textsc{Exp-S}, \textsc{Var-D}, and \textsc{Exp-D}. Here, \textsc{Var-S} denotes calculating variance in a single window, \textsc{Exp-S} represents computing expectation in a single window; \textsc{Var-D} indicates variance calculation in dual-window, and \textsc{Exp-D} signifies expectation computation in dual-window.
        
\begin{figure}[h] %htbp
            \centering
            \subfloat[]{
                    \label{fig:var_A_1}
                    \includegraphics[width=0.48\linewidth]{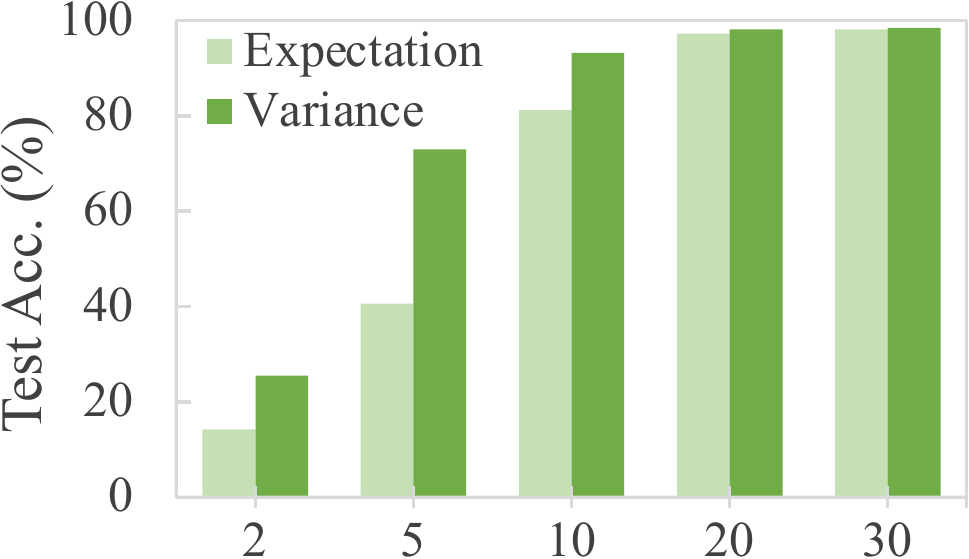}}
            \subfloat[]{
                    \label{fig:var_A_2}
                    \includegraphics[width=0.48\linewidth]{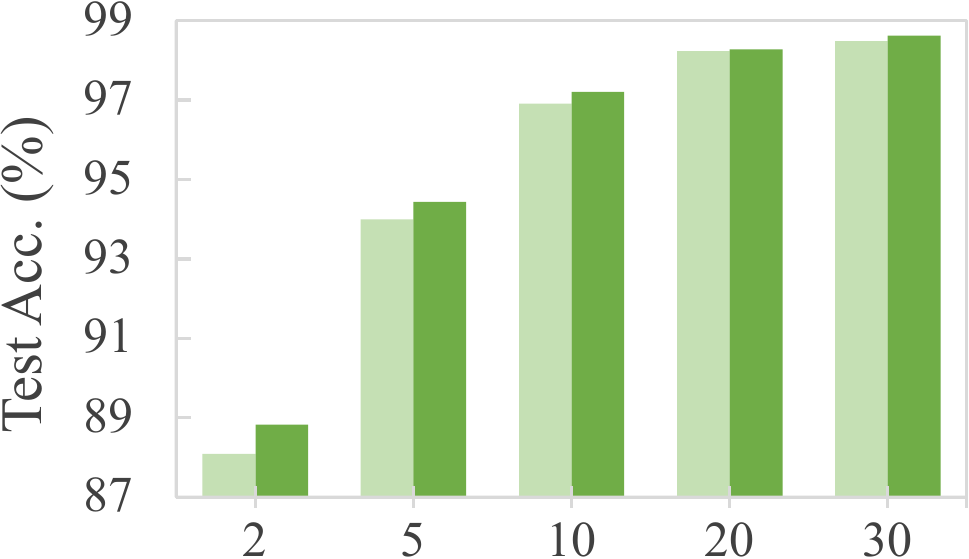}}\\
            \subfloat[]{
                    \label{fig:var_S_1}
                    \includegraphics[width=0.48\linewidth]{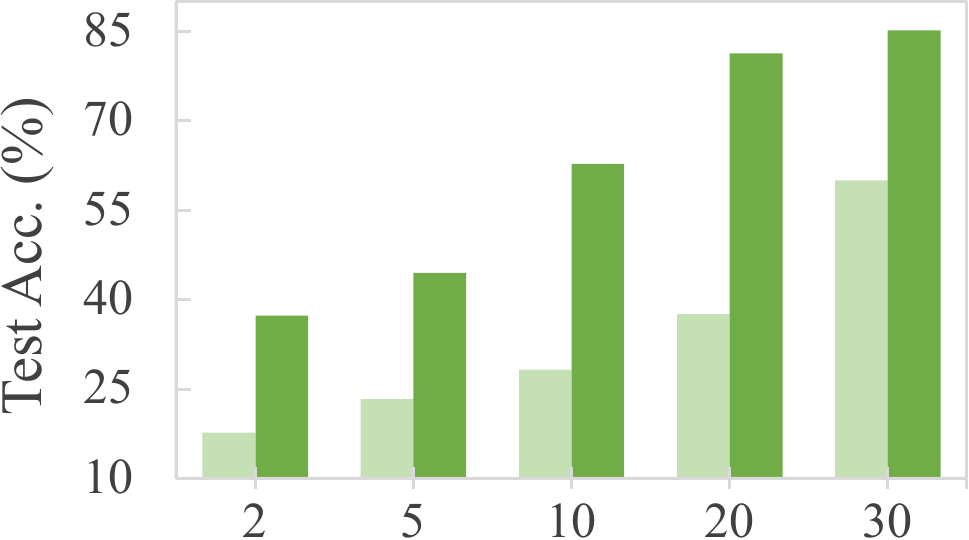}}
            \subfloat[]{
                    \label{fig:var_S_2}
                    \includegraphics[width=0.48\linewidth]{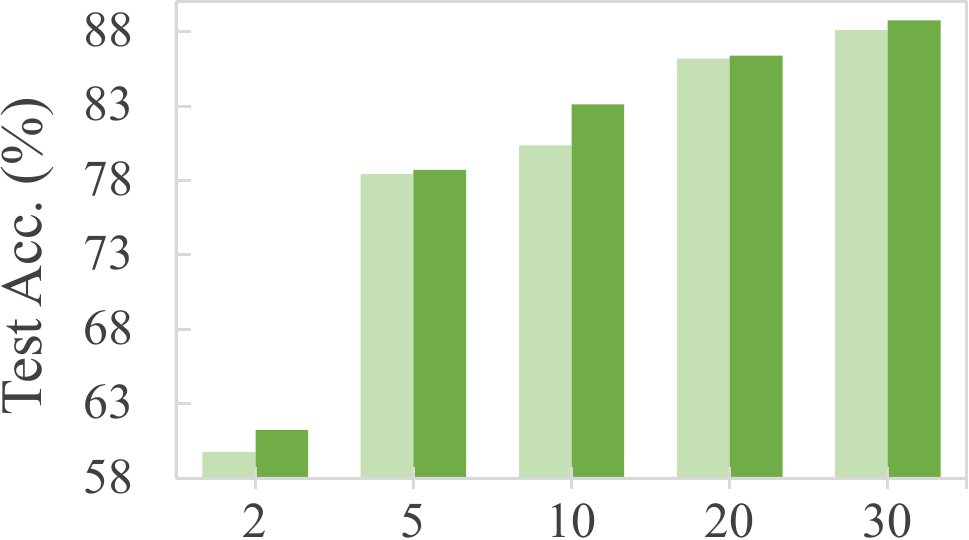}}
            \captionof{figure}{Ablation study on the summary statistics. We validated the effectiveness of variance measurement under single-window and dual-window settings on OrganAMNIST (a)(b) and OrganSMNIST (c)(d). In (a) and (c), we contrast the \textsc{Exp-S} and \textsc{Var-S} strategies within an early 10-epoch window. (b) and (d) explore the \textsc{Exp-D} and \textsc{Var-D} strategies in dual-window setting.}
            \label{fig:bar_var_a_s}
            \vspace{-10pt}
        \end{figure}

        \vspace{5pt}
        \fontsize{10}{12}\textbf{Effectiveness of Variance.}
        In this section, to demonstrate the effectiveness of variance measurement, %we first compare the results of calculating the expectation and variance of the samples' error vectors within a single window.
        we display the test accuracy results of calculating the expectation and variance of the samples' error vectors within a single window or dual windows on different datasets. 
        As shown in \cref{fig:bar_var_a_s}, these results were obtained under varied selection rates from 2\% to 30\%. 
        
        % The effectiveness of variance as a metric in coreset selection is clearly illustrated in Figure 5, which displays the test accuracy results for both OrganAMNIST and OrganSMNIST datasets under varied selection rates from 2% to 30%. 
        % \captionsetup{font=small}
        % \begin{figure}[htbp]
        %     \centering
        %     \includegraphics[width=0.4\linewidth]{paragraphs//figs/var-A-1.pdf}
        %     \includegraphics[width=0.4\linewidth]{paragraphs//figs/var-A-2.pdf}\\
        %     % \vspace{5pt}
        %     \includegraphics[width=0.4\linewidth]{paragraphs//figs/var-S}-1.pdf}
        %     \includegraphics[width=0.4\linewidth]{paragraphs//figs/var-S}-2.pdf}
        %     % \setlength{\abovecaptionskip}{-5pt}
        %     \captionof{figure}{Ablation study on the summary statistics. }
        %     \smallskip 
        %     \parbox{\linewidth}{%\raggedright 
        %     We validated the effectiveness of using variance under both single-window and dual-window settings on OrganAMNIST(a)(b) and OrganSMNIST(c)(d). In (a)(c), we compared the performance of \textsc{Exp-S} and \textsc{Var-S} strategies. In (b)(d), we compared the performance of \textsc{Exp-D} and \textsc{Var-D} strategies.
        %     }
        %     \label{fig:bar_var_a_s}
        % \end{figure}        
        % \captionsetup{font=small}
        % \vspace{-0.6cm}
        \setlength{\abovecaptionskip}{5pt}
        
        The first thing we notice is that, on both datasets, as the selection rate increases, the effectiveness of the models trained on the samples selected by both statistics tends to increase on the test set. This is intuitive because as the number of samples selected increases, the information richness of the selected samples are effectively preserved.

        Besides, we can observe that at each selection rate, the variance measurement has better performance in coreset selection compared to the expectation measurement, and this advantage is especially significant at low selection rates. 
        For example, in \cref{fig:var_S_1}, the test accuracy under \textsc{Var-S} is at least 20\% higher than under \textsc{Exp-S} for all compared selection rates.
        % the variance setting are at least 20\% higher than under the expectation setting for all selection rates in the figure.
        The consistent superiority of variance (\textsc{Var-S} and \textsc{Var-D}) suggests its robustness as a measure, further proving our previous points that (1) \textit{Expectation} may mask variability within the data by averaging contributions, thereby potentially underrepresenting the underlying fluctuations. %has inherent limitation in reflecting data fluctuations, 
        Samples with large expectation values may be consistently predicted incorrectly over many training iterations, indicating them too noisy/difficult and detrimental to the model's performance;
        (2) \textit{Variance} captures the degree to which sample contributions fluctuate over training iterations.  %, thus identifying more important samples. At low selection rates, samples with higher variance of the error across epochs imply a greater contribution to the model's generalization ability.        
        High variance in sample errors suggests that their influence on the model is not consistent but varies significantly, potentially due to their informative nature or because they are challenging for the model to learn. At low selection rates, samples with higher variance are indicative of a greater potential to contribute to the model's generalization ability, as they embody the critical challenges within the learning task.
        % sample with large \textit{Expectation}: continuously large error, model always fail to classify; 
        % with large \textit{Variance}: fluctuate, model continuously learning, effective, the most valuable samples for model training.
        % has inherent limitations: as a measure, expectation is susceptible to extreme values and therefore may not accurately reflect the importance of the sample in some cases, whereas variance is more robust and can better describe the distribution of the sample error over a period of the training dynamics.

        % It is also worth noting the almost linear improvement in test accuracy with increased selection rates when using variance. This suggests that variance is capturing the essential traits of the data distribution that correlate with performance, providing a compelling argument for its use in scenarios where the coreset needs to be both representative and concise.
        
\begin{figure}[h] %htbp
            \centering
            \subfloat[]{
                    \label{fig:dual_A_var}
                    \includegraphics[width=0.48\linewidth]{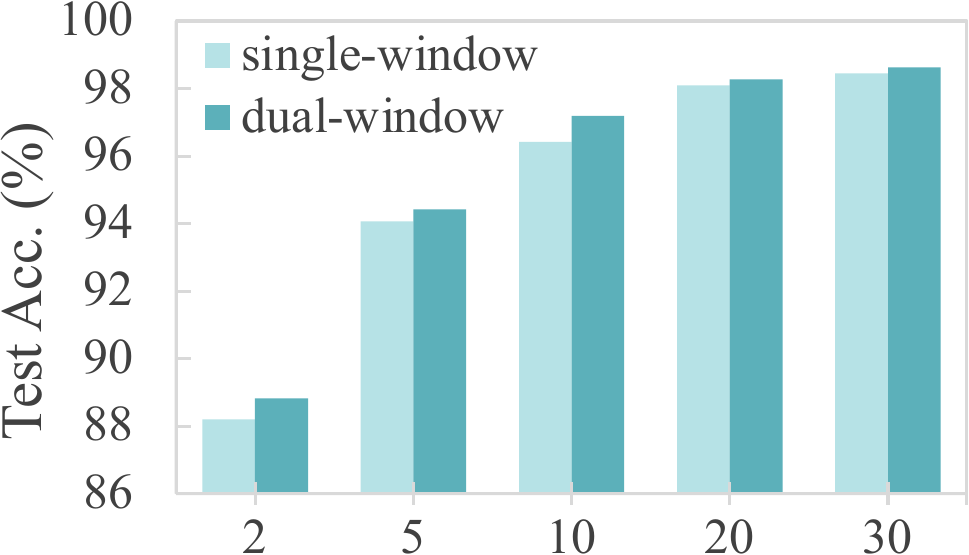}}
            \subfloat[]{
                    \label{fig:dual_A_exp}
                    \includegraphics[width=0.48\linewidth]{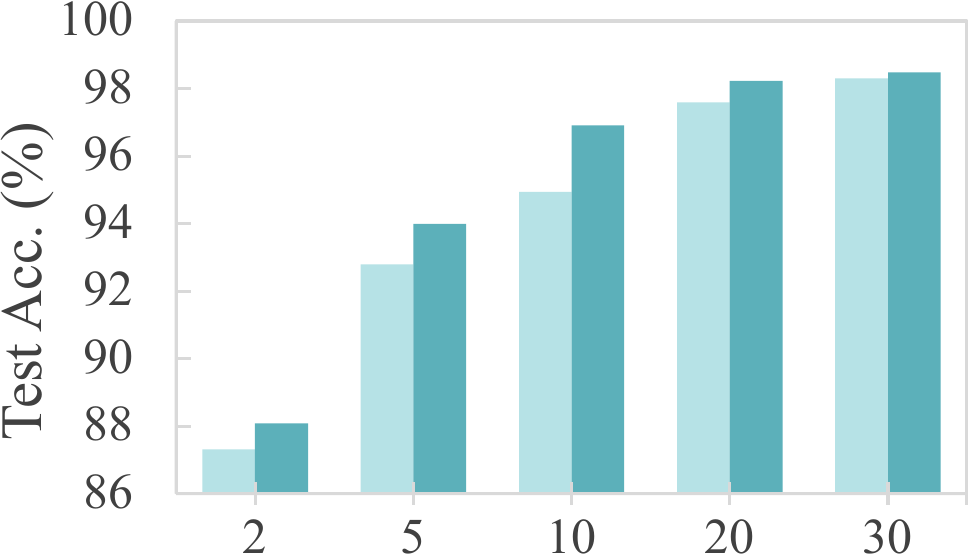}}\\
            \subfloat[]{
                    \label{fig:dual_S_var}
                    \includegraphics[width=0.48\linewidth]{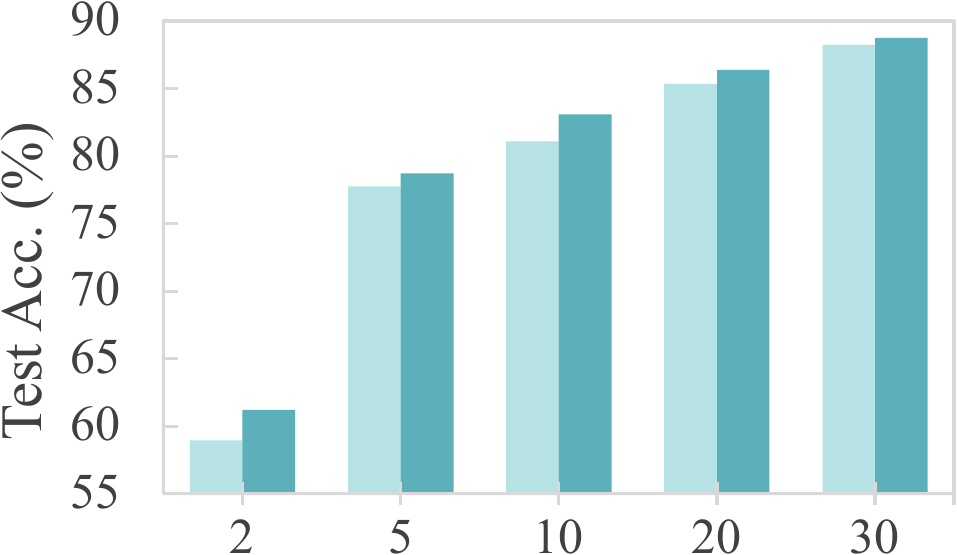}}
            \subfloat[]{
                    \label{fig:dual_S_exp}
                    \includegraphics[width=0.48\linewidth]{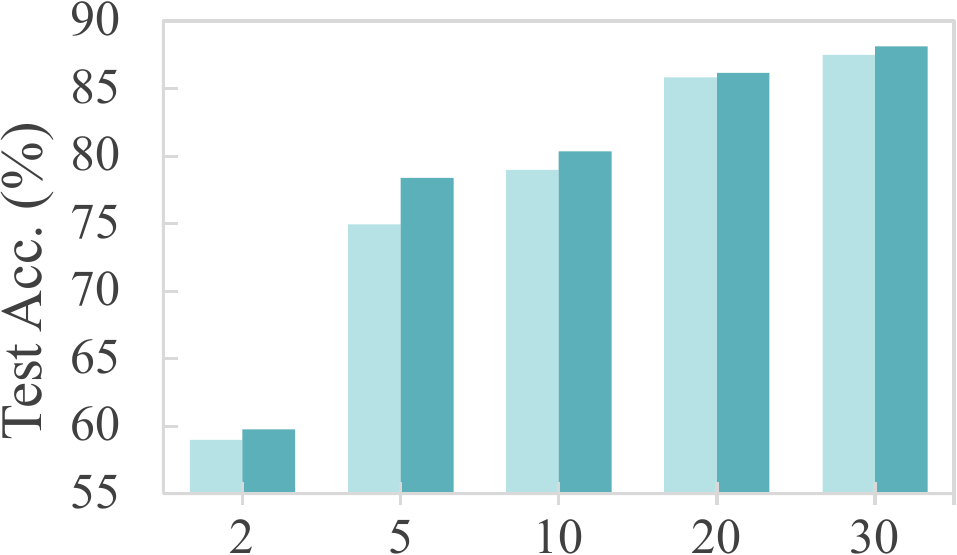}}
            \setlength{\abovecaptionskip}{3pt}
            \captionof{figure}{Ablation study on the window setting. The results are obtained on OrganAMNIST (top row) and OrganSMNIST (bottom row). Performance of the \textsc{Var-S} versus \textsc{Var-D} strategies is illustrated in (a) and (c), while (b) and (d) show comparisons between \textsc{Exp-S} and \textsc{Exp-D} strategies.}
            \label{fig:bar_dual_a_s}
            \vspace{-0.4cm}
        \end{figure}

        \vspace{5pt}
        \fontsize{10}{12}\textbf{Effectiveness of Dual-window.}
        In this section, we demonstrate the effectiveness of the dual window setting and analyze the results for different window combinations. 
        First, we compare the performance of using single-window and dual-window on different datasets (as shown in \cref{fig:bar_dual_a_s}). Similar to the former part, we utilized the variance and expectation of errors within single and dual windows as importance metrics.
        The results consistently demonstrate the advantage of dual windows over single window across all selection rates. This advantage, akin to the findings from the variance ablation experiments, is more pronounced at lower selection rates. 
        For instance, in \cref{fig:dual_S_var}, on dataset OrganSMNIST, at selection rates of 2\%, the variance calculated within dual windows exhibited an improvement of 2.29\%, compared to the single-window approach, suggesting that employing dual-window calculation for scores enables more effective capturing of the diversity and variability of sample importance.

        % \captionsetup{font=small}
        \vspace{-0.6cm}
        
        % \begin{figure*}
        %     \centering
        %     \includegraphics[width=0.99\linewidth, height=0.3\linewidth]{paragraphs//figs/line_phase_s.png}
        %     \includegraphics[width=0.99\linewidth, height=0.3\linewidth]{paragraphs//figs/line_phase_a.png}
        %     \caption{Comparison of different window combinations. These windows represent different training phases. (a)-(d) and (e)-(f) are the experimental results obtained on the OrganSMNIST and OrganAMNIST datasets, respectively. Different curves represent different window combinations (both single and dual windows).}
        %     \label{fig:line_phase}
        % \end{figure*}
        \captionsetup[subfig]{labelformat=empty}
        \setlength{\abovecaptionskip}{5pt}
        \begin{figure*}[!t]
            \centering
            \includegraphics[width=0.49\linewidth]{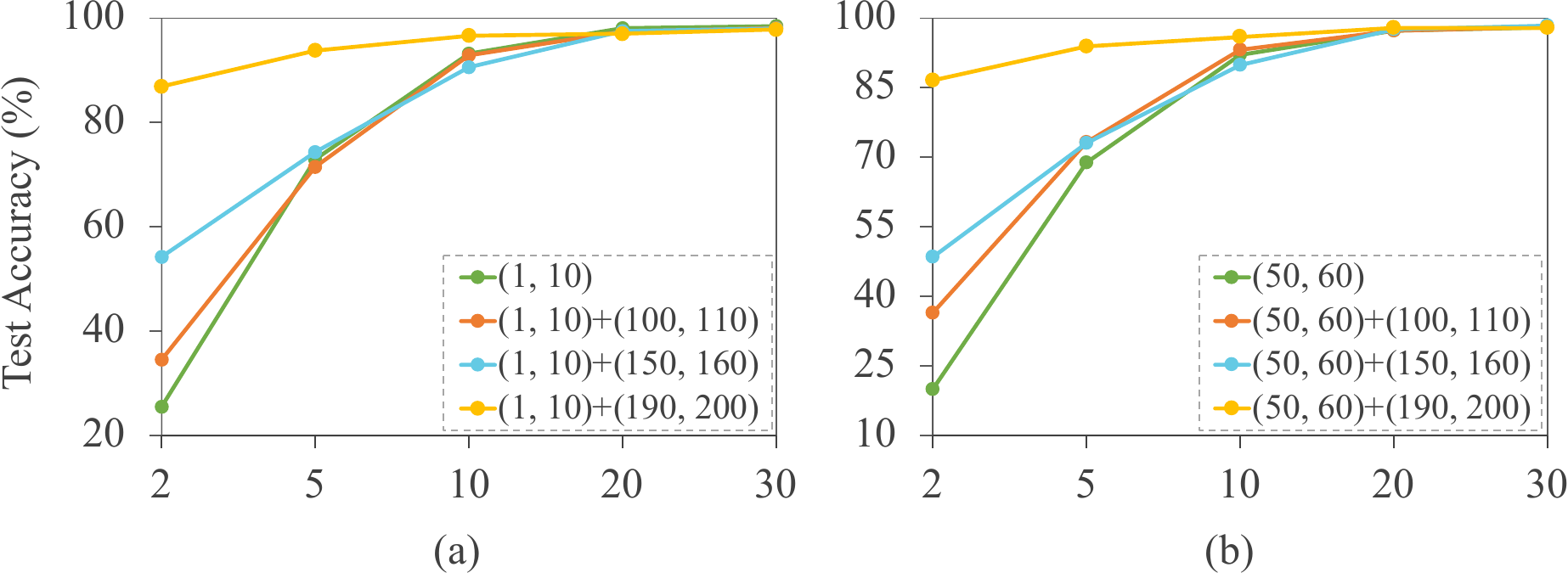} \hspace{-1mm}
            \includegraphics[width=0.49\linewidth]{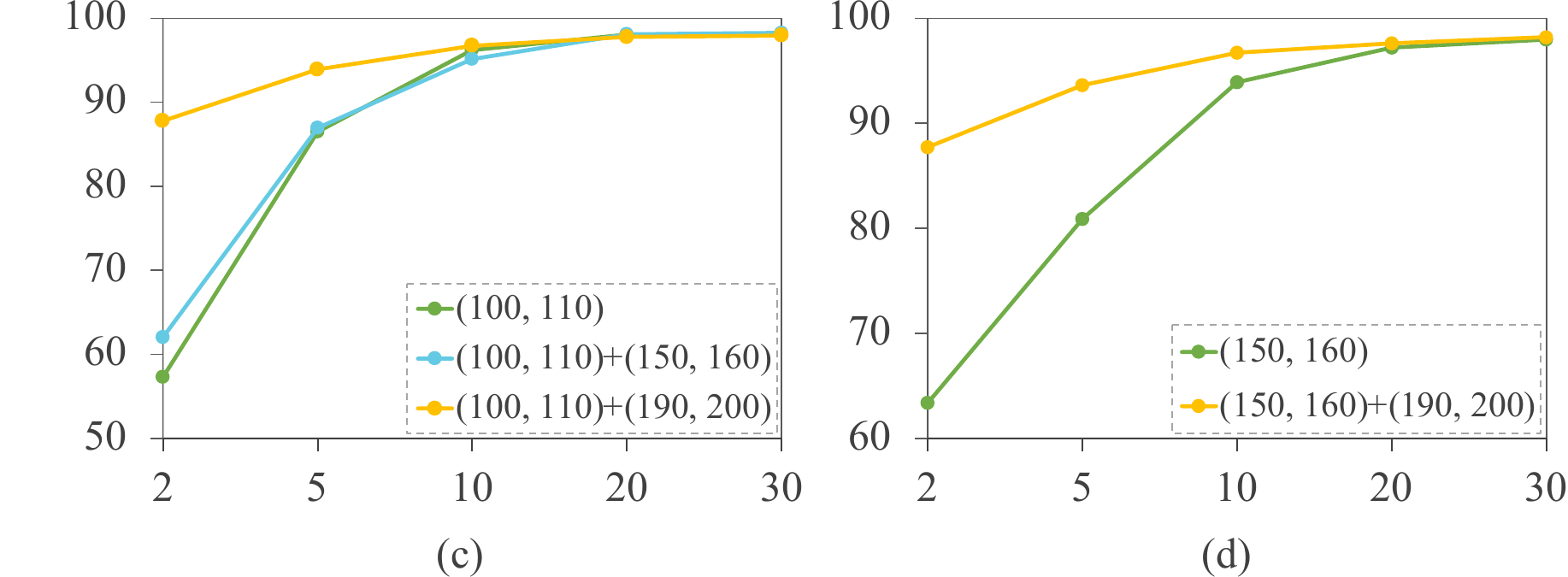} \\
            \vspace{1pt}
            \includegraphics[width=0.49\linewidth]{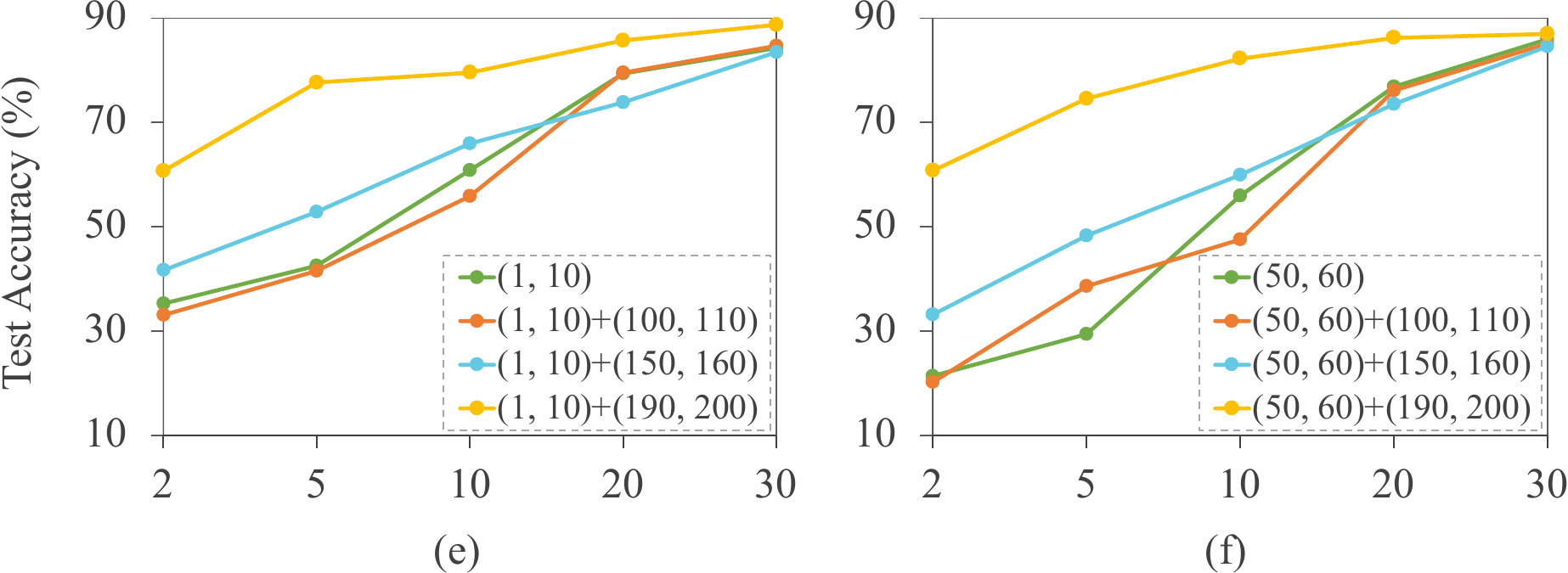} \hspace{-1mm}
            \includegraphics[width=0.49\linewidth]{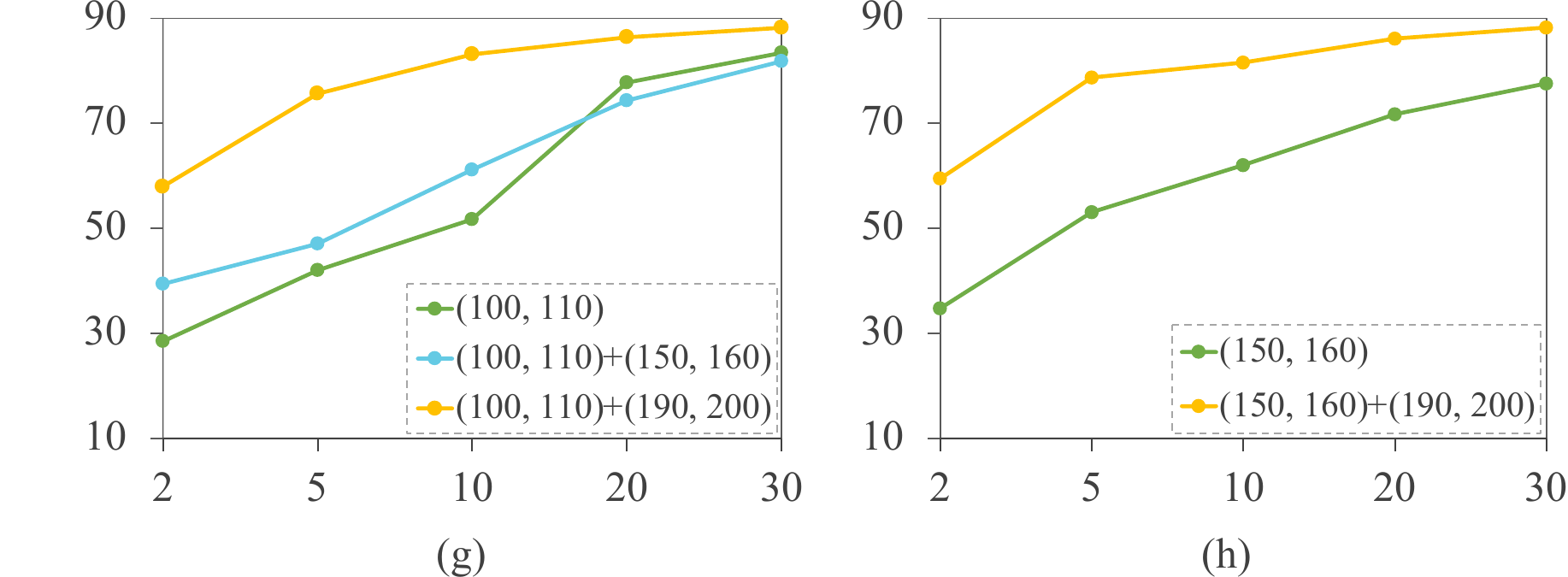}
            \caption{Comparison of different window combinations. These windows represent different training phases. 
            (a)-(d) show experimental results for OrganSMNIST, and (e)-(h) for OrganAMNIST, with each line depicting a unique window combination (single or dual windows).
            }
            %For instance, "(1, 10)" indicates a single-window phase early in training. In contrast, combinations like "(1, 10)+(100, 110)" suggest a dual-window setup where two distinct phases of training are considered.
            \label{fig:line_phase}
            \vspace{-0.6cm}
        \end{figure*}
        
        % The lower the selection ratio, the earlier epochs (low-level features).
        Moreover, in the dual-window setting, we further explore
        % the effect of the combination of windows at different periods on model performance. 
        how combining windows from different periods affects model performance. 
        \cref{fig:line_phase} reveals two critical insights: (1) At a high compression rate of 2\%, dual-window combinations show a definitive advantage over single-window ones on both datasets. This can be attributed to the dual-window's ability to encapsulate more diverse information from different training stages, providing a broader perspective for coreset selection. (2) As the selection rate increases, 
        % allowing for larger data budgets, corresponding to the need of capturing a wider range of training dynamics. 
        the data budget increases and accordingly a wider range of training dynamics need to be captured.
        The implication here is that the windows selected for the dual-window setting should ideally come from a later training stage, when the model has begun to stabilize and the samples are more reflective of the generalization capabilities required for the test. Results on OrganAMNIST suggest that early dual-window stages may not suffice for selecting a representative coreset.
        Also, please refer to the Supplementary for the comparison of the dual-window variance with the entire-process variance.
        
        %(2) Dual-window combinations do not always outperform single-window configurations. For example, on OrganAMNIST, the dual-window combination (1, 10)+(100, 110) has lower test accuracy than the single-window (1, 10). as the number of samples selected increases, we need to consider larger range of training dynamics, thus dual-window should select later epochs.
        % We use the grid search with 10-step size to find the optimal window combination ($t_e$, $t_l$) for different datasets and selection rates $\alpha$. For each dataset, we list the optimal ($t_e$, $t_l$) for every $\alpha$ as follows in the format of the tuple (($t_e$, $t_l$), $\beta$). For OrganAMNIST, the optimal setting is ((50, 100), 2\%), ((50, 100), 5\%), ((0, 100), 10\%)... . For OrganSMNIST, the optimal setting is ((0, 100), 2\%), ((50, 100), 5\%), ((0, 100), 10\%)... For CIFAR10, the optimal setting is ((0, 100), 2\%), ((50, 100), 5\%), ((0, 100), 10\%)...

%% file: paragraphs/tabs/tab_low.tex
\begin{table*}[ht]
\renewcommand{\arraystretch}{0.95}
	\caption{Performances of ResNet-18 using various coreset selection methods on MedMNIST medical datasets. % The model trained with the full dataset achieves \(98.39\%\) and \(91.76\%\) accuracy.
 All training is repeated 3 times with different random seeds to calculate mean accuracy with standard deviation.
 The first and second best results in each column are marked in red and blue, respectively. }
        \vspace{-7pt}
	\label{tab:low_SA}
	\begin{center}
		\begin{normalsize}
			\setlength{\tabcolsep}{3mm}
			\begin{tabular}{c|ccccc|ccccc}
				\hline
                    \toprule[0.8pt]
                    & \multicolumn{5}{c}{\textbf{OrganAMNIST}} &\multicolumn{5}{c}{\textbf{OrganSMNIST}}\\ 
                    % \cmidrule(r){2-7} \cmidrule(r){7-11} 
                    $\alpha$&{2\%}&{5\%}&{10\%}&{20\%}&{30\%}&{2\%}&{5\%}&{10\%}&{20\%}&{30\%}\\
                    \midrule[0.6pt]
                    \midrule[0.6pt]
                    Full dataset   &\multicolumn{5}{c}{\format{98.39}{0.02}}  &\multicolumn{5}{c}{\format{91.76}{0.55}}\\
                    \midrule[0.6pt]
                    
                    Random   &\format{87.63}{0.76} &\format{93.43}{0.65} &\format{\color{myblue}95.68}{0.45} &\format{\color{myblue}97.30}{0.13} &\format{98.14}{0.13}    &\format{\color{myblue}58.74}{0.76}  &\format{\color{myblue}73.10}{1.84} &\format{\color{myblue}80.95}{0.66} &\format{\color{myblue}85.77}{1.14} &\format{\color{myblue}87.64}{0.72} \\
                    Forgetting \cite{toneva2018empirical} &\format{15.58}{0.47} &\format{38.53}{2.78} &\format{75.85}{1.69} &\format{97.22}{0.38} &\format{98.11}{0.04}       &\format{4.33}{0.22} &\format{22.33}{0.31} &\format{33.15}{0.60} &\format{64.43}{1.23} &\format{81.28}{2.31} \\
                    Entropy \cite{coleman2019selection} &\format{41.46}{3.46} &\format{55.37}{1.4} &\format{69.04}{1.16} &\format{77.07}{1.29} &\format{91.98}{0.83}   &\format{27.93}{2.08} &\format{41.69}{0.73} &\format{59.86}{1.84} &\format{78.69}{2.13} &\format{86.20}{0.54}\\	
                    EL2N \cite{paul2021deep}  &\format{14.16}{1.14} &\format{40.68}{3.36} &\format{81.25}{3.22} &\format{97.25}{0.24} &\format{\color{myblue}98.16}{0.30}    &\format{17.63}{1.59} &\format{23.24}{1.88} &\format{28.24}{1.44} &\format{37.58}{1.53} &\format{60.06}{2.14} \\
                    AUM \cite{pleiss2020identifying} &\format{12.81}{2.62} &\format{35.10}{3.46} &\format{68.44}{0.95} &\format{93.76}{1.89}&\format{98.12}{0.14}    &\format{4.56}{0.18} &\format{7.01}{1.24} &\format{22.13}{1.86} &\format{39.87}{2.19} &\format{65.93}{1.61}  \\
                    CCS \cite{zheng2022coverage} &\format{\color{myblue}88.05}{0.62} &\format{\color{myblue}93.51}{0.10} &\format{95.58}{0.32} &\format{96.86}{0.25} &\format{97.18}{0.08}    &\format{58.43}{0.25} &\format{71.73}{0.83} &\format{78.46}{0.18} &\format{83.64}{0.55} &\format{84.94}{0.22} \\
                    \rowcolor[HTML]{EFEFEF}
                    \method (Ours)  &\format{\color{myred}88.83}{0.88} &\format{\color{myred}94.43}{1.32} &\format{\color{myred}97.20}{0.34} &\format{\color{myred}98.27}{0.57} &\format{\color{myred}98.63}{0.34}     &\format{\color{myred}61.23}{0.75} &\format{\color{myred}78.71}{0.93} &\format{\color{myred}83.11}{0.72} &\format{\color{myred}86.38}{1.02} &\format{\color{myred}88.77}{0.43} \\
                    
                    \bottomrule[0.8pt]
				\hline
			\end{tabular}
		\end{normalsize}
  \end{center}
\end{table*}

%% file: paragraphs/related_work.tex
\section{Related Works}
    \subsection {Medical Imaging}\label{subsec:medical}
        \fontsize{10}{12}\textbf{Challenges in Medical Imaging with Deep Learning.}\ \ Medical imaging technology has brought transformative advancements to the diagnosis of a variety of diseases in the past few decades, enabling earlier detection and the development of more personalized treatment plans. 
        Deep learning (DL), in particular, has been widely used in various medical imaging tasks and has achieved remarkable success in many medical imaging applications \textcolor{myred}{\cite{panayides2020ai, zhou2018unet++, chen2018drinet, saraf2020deep, zhou2021review, dai2024unichest, HOLSTE2024103224}}, enhancing the accuracy of diagnoses through the innovative use of historical data \textcolor{myred}{\cite{litjens2017survey}}.
        
\begin{figure}[h]
            \centering
            \subfloat[]{
                    \label{fig:medical_1_1}
                    \includegraphics[width=0.3\linewidth]{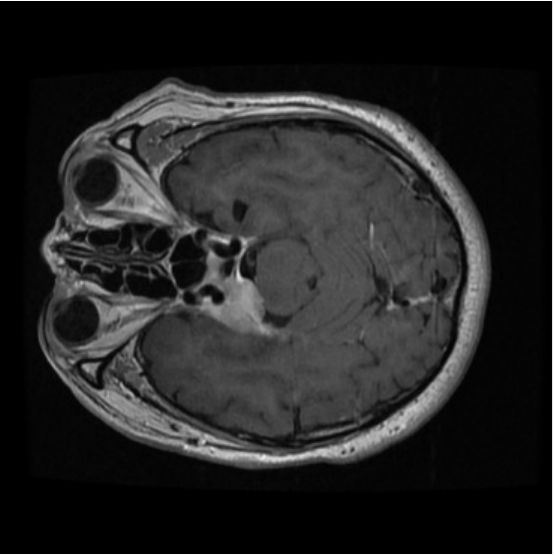}}
            \subfloat[]{
                    \label{fig:medical_3_1}
                    \includegraphics[width=0.3\linewidth]{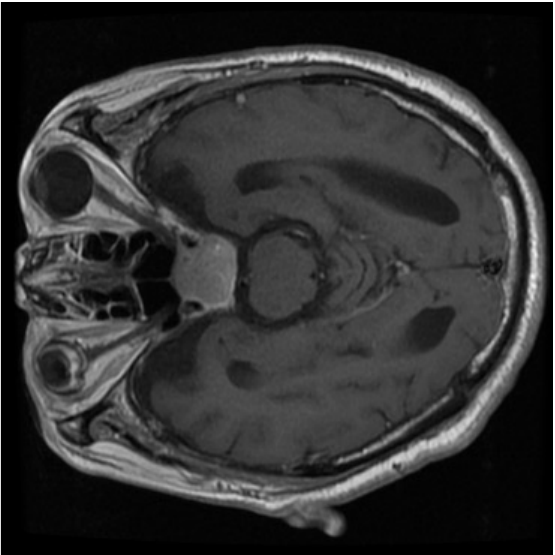}}
            \subfloat[]{
                    \label{fig:medical_2_1}
                    \includegraphics[width=0.3\linewidth]{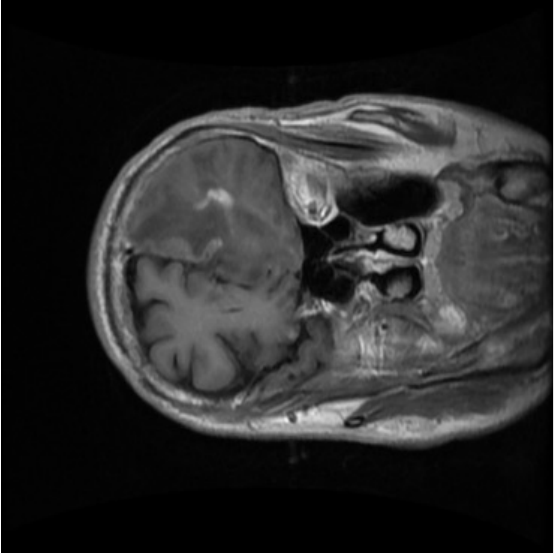}}\\
            \vspace{-5pt}
            \subfloat[]{
                    \label{fig:medical_1_2}
                    \includegraphics[width=0.3\linewidth]{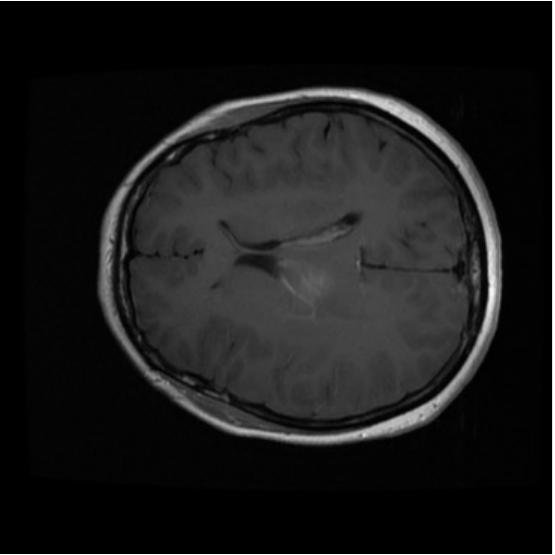}}
            \subfloat[]{
                    \label{fig:medical_3_2}
                    \includegraphics[width=0.3\linewidth]{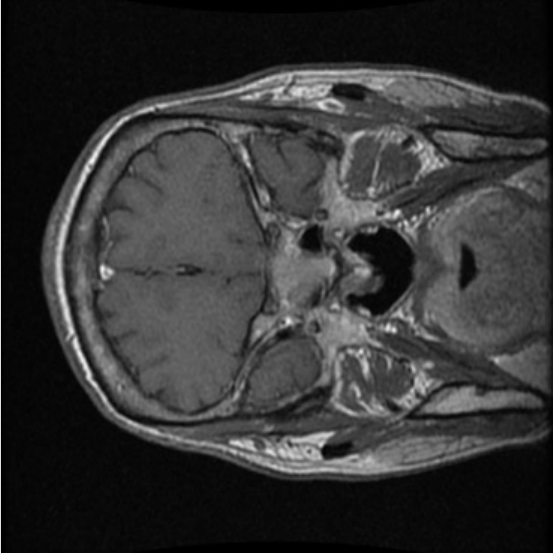}}
            \subfloat[]{
                    \label{fig:medical_2_2}
                    \includegraphics[width=0.3\linewidth]{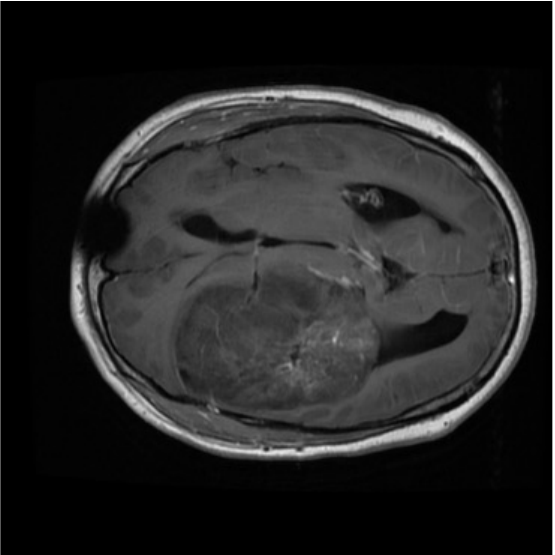}}
            \setlength{\abovecaptionskip}{3pt}
            % \end{minipage}
            % \captionsetup{labelformat=empty}
            \captionof{figure}{Examples of the intra-class variation and inter-class similarity in medical image classification. These axial brain tumor images come from the public dataset provided by Jun Cheng et al.\textcolor{myred}{\cite{cheng2017brain}}. Each column respectively represents a brain tumor category: meningioma (a)(d), pituitary (b)(e), and glioma (c)(f). 
            The variation within the same category can be noticed by observing the two instances in each column.
            Furthermore, the similarity between different classes is illustrated by comparing (a)(b), (c)(e), and (d)(f).}
            \label{fig:medical}
            \vspace{-0.4cm}
        \end{figure}

        Despite substantial progress, integrating DL into medical imaging is fraught with challenges \textcolor{myred}{\cite{ker2017deep}}. Its effectiveness is largely dependent on the availability of large, well-annotated datasets tailored for specific tasks and reliant on advances in high-performance computing. The necessity for vast complex datasets introduces complications such as 
        data quality inconsistencies from different imaging equipment and protocols. 
        % inconsistencies in data quality, arising from variations in imaging equipment and protocols. 
        Moreover, the extensive volume of medical data demands significant computational resources, posing logistical challenges for efficient processing \textcolor{myred}{\cite{zhou2021review}}.
        Additionally, the inherent heterogeneity of medical images, characterized by a multimodal probability distribution, complicates the model training process by requiring algorithms capable of handling diverse visual features and patterns within the data. 
        Another issue is the inter-class similarity and intra-class variation, as depicted in \cref{fig:medical}, where different diseases may appear similar, and the same disease may present differently across patients.

        \vspace{5pt}
        \fontsize{10}{12}\textbf{MedMNIST: A Standardized Dataset for Biomedical Imaging.}\ \ 
        To address some of these challenges, MedMNIST, a large-scale MNIST-like collection of standardized biomedical images, provides a comprehensive dataset for research and application. This dataset includes 12 datasets for 2D imaging and 6 for 3D, all pre-processed into 28x28 or 28x28x28 pixels with corresponding classification labels. 
        % MedMNIST encompasses primary data modalities in biomedical imaging, including abdominal CT, chest X-ray, breast ultrasound, and blood cell microscopy, making it an ideal choice for multi-modal machine learning in medical image analysis.
        Covering primary biomedical imaging modalities like abdominal CT, chest X-ray, breast ultrasound, and blood cell microscopy, MedMNIST is ideal for multi-modal machine learning in medical image analysis. 
        Additionally, it supports various classification tasks such as binary/multi-class, ordinal regression, and multi-label classification, further establishing its utility for developing and testing deep learning models in medical imaging.

        % \captionsetup{font=small}
        
    \vspace{-0.2cm}
    \subsection{Dataset Compression}\label{subsec:compression}
    The proliferation of large-scale datasets in deep learning necessitates the compression of data size to meet specific requirements, such as computational efficiency and storage constraints. % Dataset compression techniques, such as dataset pruning and distillation, play a crucial role in addressing these challenges.
    Therefore, the identification of key samples serves a fundamental role not only in dataset pruning but also across a spectrum of machine learning tasks, such as active learning \textcolor{myred}{\cite{bordes2005fast, ash2019deep, emam2021active}}, where the model is trained iteratively on a subset of the dataset, and only the most informative samples are selected for inclusion in subsequent training rounds. Techniques such as uncertainty sampling and query-by-committee have been proposed to select data samples that are most beneficial for model improvement.
    Continual learning \textcolor{myred}{\cite{yoon2021online}}, where a memory buffer is maintained to store informative training samples from previous tasks for rehearsal in future tasks.
    And other problems like noisy learning \textcolor{myred}{\cite{mirzasoleiman2020coresets}}, clustering \textcolor{myred}{\cite{bateni2014distributed}}, imbalanced learning \textcolor{myred}{\cite{cui2021parametric, hong2023long, hong2024on}}, semi-supervised learning \textcolor{myred}{\cite{borsos2021semi}}, and unsupervised learning \textcolor{myred}{\cite{chhaya2020coresets}}. 
    
    Dataset pruning, or coreset selection, can generally be categorized into several groups: 
    Score-based techniques \textcolor{myred}{\cite{coleman2019selection, feldman2020neural, sorscher2022beyond, toneva2018empirical, paul2021deep, meding2021trivial}}, methods driven by coverage or diversity considerations \textcolor{myred}{\cite{xia2022moderate, welling2009herding, sener2017active}}, 
    and strategies grounded in optimization \textcolor{myred}{\cite{yang2205dataset, mirzasoleiman2020coresets, killamsetty2021grad, killamsetty2021glister, killamsetty2021retrieve, wei2015submodularity, kaushal2021prism, kothawade2021similar, hong2024diversified}}.
    Specifically, score-based techniques first assign an importance score to each training sample based on its influence over a specific permanence metric during model training. The samples are then sorted by their scores, and those within a certain range are selected to construct the coreset.

    Besides, in the sphere of data-efficient deep learning, 
    % associated topics include 
    techniques like data distillation \textcolor{myred}{\cite{cazenavette2022dataset, liu2022dataset}} and data condensation \textcolor{myred}{\cite{kim2022dataset, liu2022dataset, cui2022dc}} 
    % , which seeks to condense the knowledge contained in a large dataset into a smaller, distilled dataset.
    aim to condense knowledge from large datasets into smaller, distilled ones.
    This often involves training a smaller "student" model to mimic % the behavior of 
    a larger "teacher" model, effectively transferring the knowledge from the larger dataset to the distilled one. 
    Most distillation methods are evaluated on natural image datasets and their effectiveness lack comprehensive verification on medical datasets. To the best of our knowledge, a recent work \textcolor{myred}{\cite{yu2024progressive}} proposed a comprehensive benchmark for evaluating the medical image dataset distillation.
    \vspace{-0.2cm} % before Sec.6

%% file: paragraphs/limitation_future.tex
\section{Limitation \& Future Work}
    % 1. explore different compression limits for different dataset (balance between acc & efficiency) 
    While our EVA coreset selection strategy demonstrates superior performance in high compression scenarios, as evidenced by the comparative analysis presented in \cref{tab:low_SA}, it's important to acknowledge the limitations that the level of accuracy achieved in scenarios demanding extreme compression may not fully meet the rigorous standards necessary for medical diagnostics. 
    Medical imaging tasks often require the highest degree of precision due to their direct impact on patient care, and there remains room for improvement in ensuring that the selected coresets are not only statistically representative but also clinically relevant.
    
    % 2. 我们没有结合 different modality data，比如在智慧医疗诊断系统中: medical image，问诊描述（医疗问答）。 
    Additionally, our current approach does not incorporate data from different modalities, which is essential in smart healthcare diagnostic systems. Such systems typically combine various types of data, including medical images, electronic health records (EHRs), patient interview descriptions, and pathology reports, for holistic analysis to enhance diagnostic accuracy.
    
    % especially in the trend of LLM application...
    Future research could explore different compression limits for various datasets to find the optimal balance between accuracy and efficiency. This involves systematically determining how much data can be pruned while maintaining sufficient performance for clinical applications.
    Moreover, %there is a promising avenue to extend our work by integrating multimodal data, 
    integrating multimodal data is a promising avenue to extend our work, which aligning with trends in applying large language models (LLMs) and other advanced AI techniques in healthcare. Such integration could enhance the robustness and applicability of our coreset selection strategy, particularly in systems requiring diverse data synthesis for effective decision-making.

    % Future efforts could also explore adaptive coreset selection mechanisms that dynamically adjust based on the specific characteristics of the data being processed. This could help in managing the inherent variability in medical datasets and tailor the pruning process to the needs of specific medical tasks.

    % In conclusion, while our current methodology offers significant advancements in coreset selection for medical imaging, the integration of multimodal data and the exploration of adaptive strategies represent critical steps forward for the field.

    % Another limitation of our approach is its generalizability to different types of medical data and imaging modalities. The performance of the EVA strategy has been validated on specific datasets, but medical images can vary widely in their characteristics, which may impact the strategy's effectiveness.
        
    % Future work could focus on exploring the applicability of our approach to other types of medical data including different data modalities, 
    
    % and investigating the integration of our coreset selection strategy with other machine learning tasks such as noisy learning and continuous learning. 
    % Additionally, further research could aim to optimize the dual-window approach and variance measurement technique to enhance their effectiveness in different medical imaging contexts.
    

%% file: paragraphs/conclusion.tex
\vspace{-0.5cm}
\section{Conclusion}
    In this paper, we identify and analyze the limitations of existing coreset selection methods in capturing the evolutionary nature of model training and the fluctuations in sample importance within medical image datasets. To address this challenge, we introduced a novel sample scoring strategy, \Methods, which incorporates a dual-window method to consider the training dynamics at different stages and employs a variance measurement of samples' error vectors for a more precise evaluation of sample importance.
    Extensive evaluations on various datasets and neural network 
    architectures demonstrate the superior performance of our proposed \method strategy.
    
% The proposed \Methods strategy represents a significant advancement in the field of dataset compression for medical imaging. By enabling more efficient processing and analysis of medical image datasets, it holds promise for improving diagnostic accuracy and facilitating timely and personalized treatment plans in the medical field.

%% file: supplementary/baseline.tex
\section{Comparison Methods}\label{sec:experiment_setting}
    % \vspace{5pt}
    % \setlength{\parindent}{2em} % indent 1st line of each para
    \noindent
    (1) \textbf{Random}: Construct a coreset consisting of examples chosen from the full training set by uniform random sampling. \\
    (2) \textbf{Forgetting}: Construct a coreset composed of examples with the highest forgetting scores. The forgetting score counts how many times the forgetting happens during model training, i.e. an example was misclassified in the current epoch after being correctly classified in the previous epoch.  \\
    (3) \textbf{Entropy}: Construct a coreset of examples with the highest entropy score. It is an uncertainty-based method. Entropy indicates the uncertainty of a sample given a certain classifier and training epoch, and examples with higher entropy are more important for model training. \\
    (4) \textbf{EL2N}: Construct a coreset of examples with the highest EL2N score. As an approximation of the GraNd score, which measures the average contribution of each sample to the decline of the training loss at early epochs across several independent runs, the EL2N score measures the data difficulty or importance by the L2 norm of error vectors. \\
    (5) \textbf{Area under the margin (AUM)}: Construct a coreset consisting of examples with the lowest AUM score. AUM is a data difficulty and importance metric that identifies noisy and mislabeled data by observing a network’s training dynamics. (It measures the probability gap between the target class and the next largest class across all training epochs.) \\
    (6) \textbf{Coverage-Centric Coreset Selection (CCS)}: CCS jointly considers overall data coverage across a distribution and the importance of individual examples by employing a modified stratified sampling technique. In our experiments, AUM is used as the metric for determining the importance within the CCS framework.

%% file: supplementary/high.tex
\section{Performance at High Selecion Rates}
    We provide more comparison results between \method and other SOTA baselines at high selection rates.
    As reported in \cref{tab:high_SA}, \method consistently exhibits superior performance in the majority of cases. Notably, \method outperforms the full dataset at high selection rates, for instance, it achieves 98.80\% accuracy using only half of the OrganAMNIST data, compared to the 98.39\% accuracy with the full dataset, underlining its capability to maintain or even enhance model performance despite utilizing a pruned dataset.
    \input{supplementary/tab_high}

%% file: supplementary/tab_high.tex
\begin{table}[htbp]
    \centering
    \caption{High selection rate performance on OrganAMNIST and OrganSMNIST with ResNet-18. The models trained with the full datasets achieves 98.39\% and 91.76\%, respectively. The first and second best results in each column are marked in \textcolor{myred}{red} and \textcolor{myblue}{blue}, respectively. %All training is repeated 3 times with different random seeds to calculate mean accuracy with standard deviation.
    }
    % \vspace{-5pt} % vertical space before the table 
    \resizebox{1\linewidth}{!}{ 
    \begin{large}
    % \begin{tabular}{p{2cm}<{\centering}|ccc|ccc}
    \begin{tabular}{c|ccc|ccc}
        \toprule[1.2pt]
        
        & \multicolumn{3}{c}{\textbf{OrganAMNIST}} &\multicolumn{3}{c}{\textbf{OrganSMNIST}} \\ 
        $\alpha$  & 50\% & 70\% & 90\% & 50\% & 70\% & 90\% \\
        \midrule[0.6pt]
        \midrule[0.6pt]
        
        Random      & \format{98.14}{0.04}   & \format{98.29}{0.15}   & \format{98.44}{0.43}   & \format{89.63}{0.68}   & \format{90.23}{0.72}   & \format{91.16}{0.23} 
        \\
        Forgetting   & \color{myblue}\format{98.46}{0.16}   & \format{98.31}{0.82}   & \color{myred}\format{98.70}{0.44}   & \color{myblue}\format{91.18}{0.69}   & \color{myblue}\format{91.46}{0.38}   & \color{myblue}\format{91.58}{0.26} 
        \\
        Entropy   & \format{97.93}{0.44}   & \format{98.32}{0.03}   & \format{98.50}{0.55}   & \format{90.41}{0.27}   & \format{91.31}{0.50}   & \format{91.36}{0.78} 
        \\
        EL2N    & \format{98.39}{0.89}   & \format{98.24}{0.04}   & \format{98.48}{1.27}   & \format{89.94}{0.57}   & \format{90.43}{0.90}   & \format{91.53}{0.61} 
        \\
        AUM      & \format{98.13}{0.02}   & \color{myblue}\format{98.54}{0.76}   & \format{98.55}{0.51}   & \format{89.81}{0.52}   & \format{90.97}{0.87}   & \color{myblue}\format{91.58}{0.03} 
        \\
        CCS      & \format{97.55}{0.19}   & \format{97.56}{0.60}   & \format{97.07}{0.81}   & \format{85.53}{0.11}   & \format{85.69}{0.15}   & \format{86.38}{0.91}  
        \\
        % \rowcolor{blue!20}
        \midrule[0.6pt]
        \rowcolor[HTML]{EFEFEF} 
        \textbf{\method(Ours)}   & \textbf{\color{myred}\format{98.80}{0.56}}   & \textbf{\color{myred}\format{98.96}{0.73}}   & \textbf{\color{myblue}\format{98.68}{1.45}}  & \textbf{\color{myred}\format{91.21}{0.98}}   & \textbf{\color{myred}\format{91.94}{0.74}}   & \textbf{\color{myred}\format{91.89}{0.96}} \\
        \bottomrule[1.2pt]
    
    \end{tabular}
    \end{large}
    }
    % \vspace{-15pt}
    \label{tab:high_SA}
\end{table}

%% file: supplementary/small.tex
\section{Potential for Overfitting on Small Datasets}
    % We conduct experiments on smaller datasets with RN-18 to investigate the potential for overfitting, and \method exhibit promising performance even at high selection rates (\cref{tab:small}). DermaMNIST and PneumoniaMNIST comprises 10,015 and 5,856 images, respectively.
    % Although our method is primarily designed for large-scale datasets, it has shown promising results on relatively smaller datasets as well (\cref{tab:small}). 
    % DermaMNIST contains dermatoscope 10,015 images and PneumoniaMNIST comprises 5,856 chest X-ray images.
    In order to investigate the potential for overfitting on smaller datasets, we conduct experiments on DermaMNIST and PneumoniaMNIST (containing 10,015 and 5,856 images, respectively) using ResNet-18. As shown in \cref{tab:small}, \method exhibit promising performance compared to other baselines. 
    EVA’s dual-window approach effectively captures training dynamics, mitigating overfitting risks. 
    For extremely small datasets, pruning may not be necessary. 
    % overfitting could become an issue. However, in such cases, 
    % the need for pruning might be minimal,
    
    % as their size doesn't justify coreset selection. 
    % For extremely small datasets with high pruning rates, overfitting could become an issue. However, in such cases, pruning might not be necessary, as the dataset size itself may not justify the need for coreset selection.
    \input{supplementary/tab_small}

%% file: supplementary/tab_small.tex
% problem: resizebox, rowcolor
% \vspace{10pt}
\begin{table}[htbp]
    % \vspace{-2.0em}
    % \vspace{-5pt}
    \centering
    % \captionsetup{font={small,bf,stretch=1}, justification=raggedright}
    \caption{
    % Performances on small datasets (full: 77.73\% and 97.27\%)
    Performances on small datasets with RN-18.
    The models trained with the full datasets achieves 77.73\% and 97.27\%, respectively.
    %The first and second best results in each column are marked in \textcolor{myred}{red} and \textcolor{myblue}{blue}, respectively. % All training is repeated 3 times with different random seeds to calculate mean accuracy with standard deviation.
    }
    % \vspace{-5pt} % vertical space before the table 
    \resizebox{1\linewidth}{!}{ 
    \begin{small}
    \begin{tabular}{p{1.1cm}<{\centering}|cccc|cccc}
        \toprule[1.0pt]
        
        & \multicolumn{4}{c}{{\textbf{DermaMNIST}}} &\multicolumn{4}{c}{{\textbf{PneumoniaMNIST}}} \\ 
        $\alpha$   & 10\% & 30\% & 50\% & 70\%   & 10\% & 30\% & 50\% & 70\% \\
        \midrule[0.5pt]
        \midrule[0.5pt]
    
        Random    & 67.97   & 69.92   & 72.66   & 73.44  & \color{myblue} 92.77  & 95.70    & 95.90   & 96.48   \\
        Forgetting & 36.72   & 50.98   & 70.51   & 73.44  & 55.86   & \color{myblue} 96.48  & 95.70   & 95.51   \\
        Entropy   & 55.66   & \color{myblue} 70.31 & 73.24   & 71.48  & 71.48   & 96.09   & 95.70   & \color{myblue} 97.07 \\
        EL2N      & 10.55   & 19.73   & 55.27   & 74.41  & 26.17   & 94.34   & \color{myblue} 97.07   & 95.90    \\
        AUM       & 12.11  & 20.7   & 66.8   & \color{myblue} 75.00   & 38.87   & 96.29   & 96.29   & 96.68   \\
        CCS       & \color{myred}70.12 & \color{myred}72.27 & \color{myblue} 74.02 & 74.80   & 91.02   & 91.21   & 92.97   & 91.02   \\
    
        \rowcolor[HTML]{EFEFEF}
        \textbf{\method}  & \textbf{\color{myblue} 69.92} & \textbf{\color{myred}72.27} & \textbf{\color{myred}75}    & \textbf{\color{myred}75.2}  & \textbf{\color{myred}94.34}   & \textbf{\color{myred}97.07}   & \textbf{\color{myred}97.46}   & \textbf{\color{myred}97.66} \\
        \bottomrule[1.0pt]
    
    \end{tabular}
    \end{small}
    }
    % \vspace{-5pt}
    % \vspace{-1em}
    \label{tab:small}
\end{table}

%% file: supplementary/cifar.tex
\section{Effectiveness on Natural Image Dataset}
    The applicability of \method extends beyond medical imagery, as evidenced by our exploration of its effectiveness on natural image datasets such as CIFAR-10 and CIFAR-100. As detailed in \cref{tab:cifar10} and \cref{tab:cifar100}, our method demonstrates robust performance across varying selection rates.
    On CIFAR-10, \method attains 90.50\% accuracy at 30\% selection rate, closely approaching the full dataset's accuracy benchmark of 93.06\%. 
    In the more complex CIFAR-100 dataset, \method achieves commendable results at low selection rates, i.e., reaching 62.93\% accuracy with 30\% of the full dataset. 
    In addition, \method's performance consistently outpaces various SOTA baselines across different low selection rates tested, showcasing its generalizability and the robustness of its coreset selection efficacy in diverse image contexts.

\input{supplementary/tab_cifar}

    \captionsetup[subfig]{labelformat=empty}
    \setlength{\abovecaptionskip}{12pt}
    \begin{figure*}[htbp]
        \centering
        \includegraphics[width=0.243\linewidth]{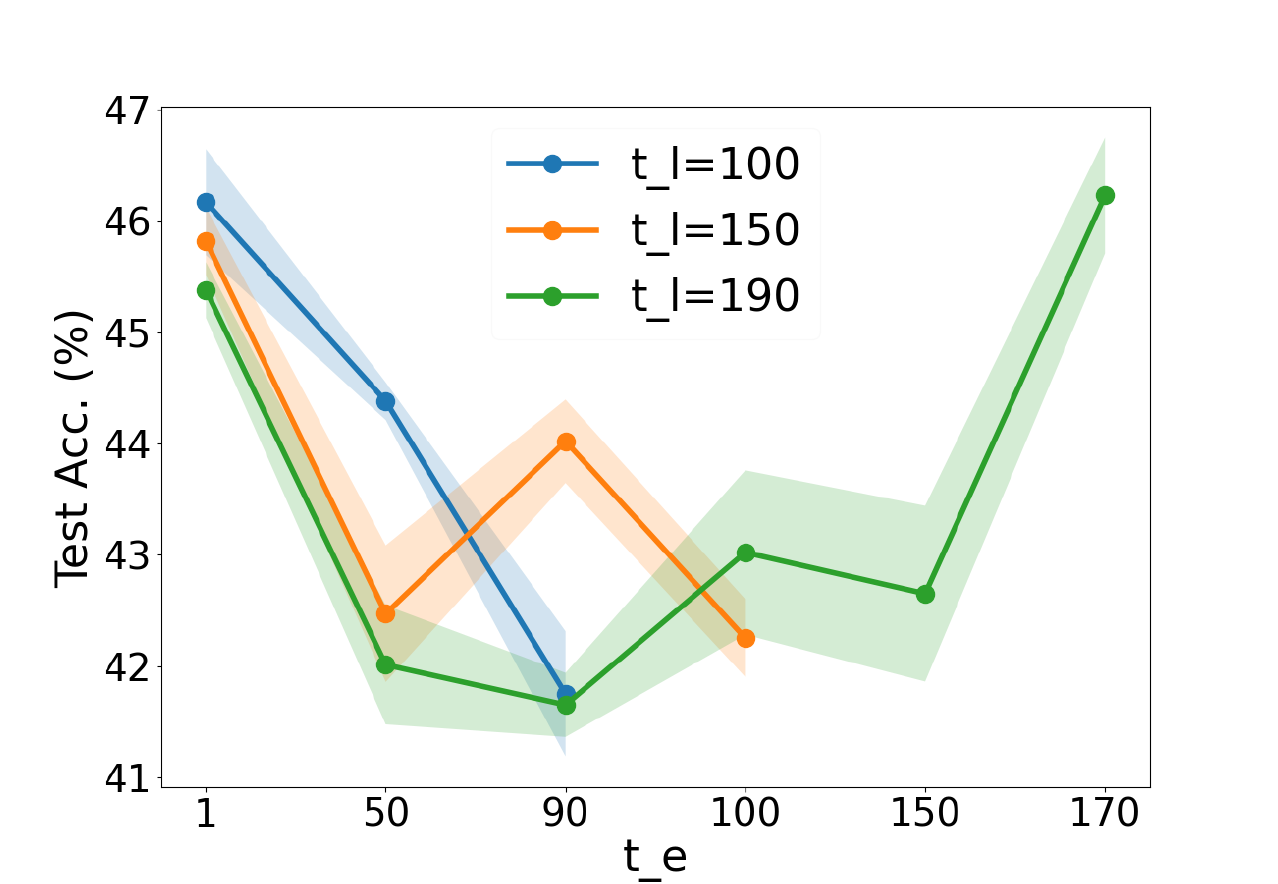} 
        \includegraphics[width=0.243\linewidth]{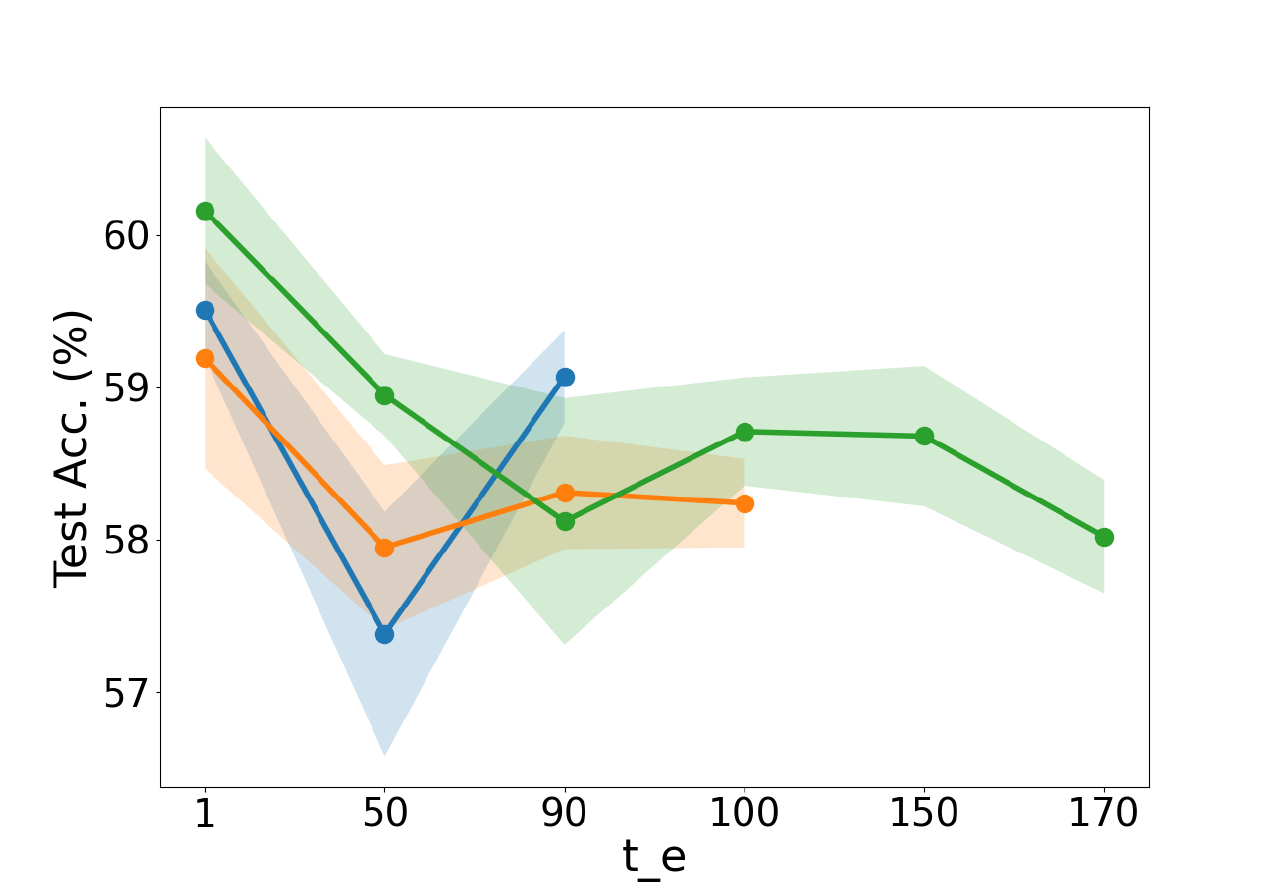} 
        \includegraphics[width=0.243\linewidth]{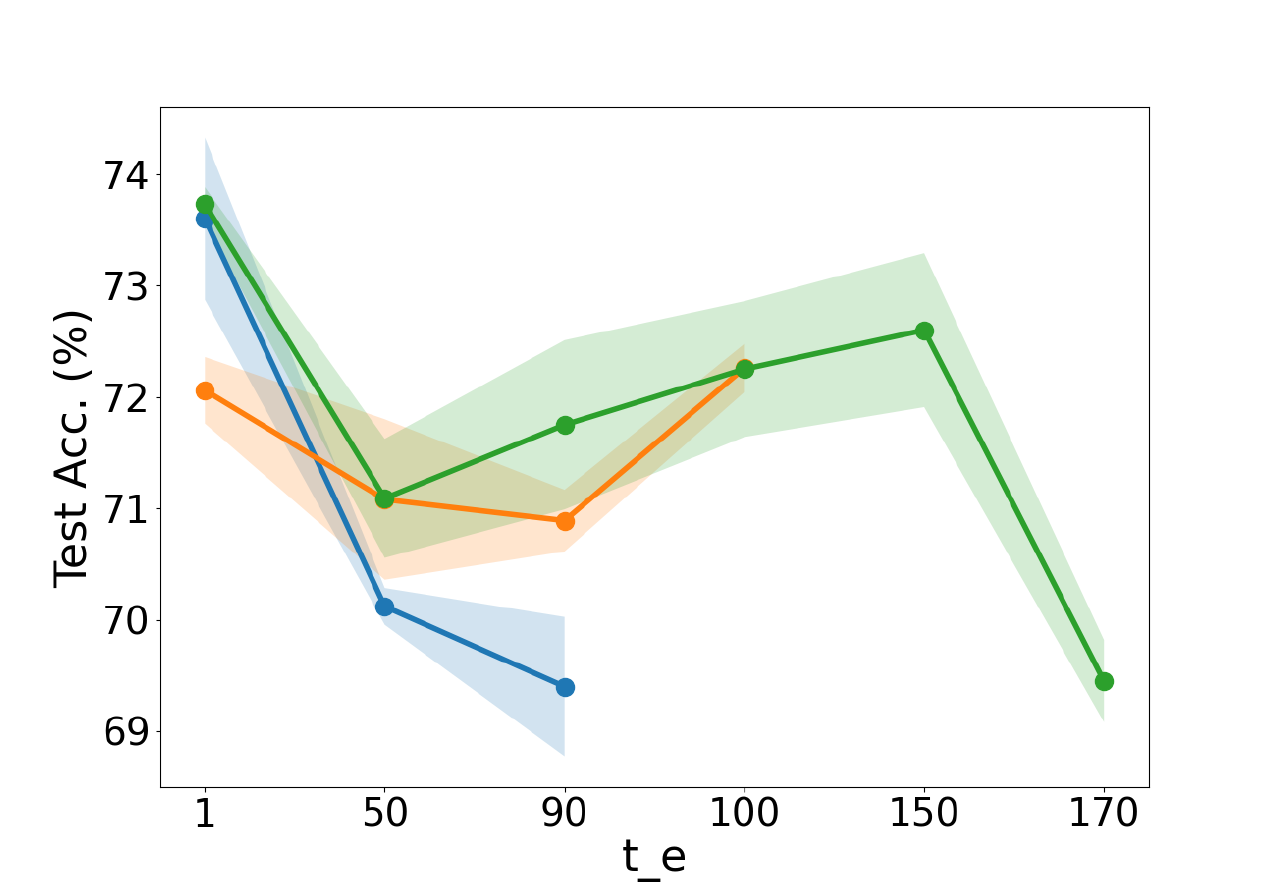} %\hspace{-1mm}
        \includegraphics[width=0.243\linewidth]{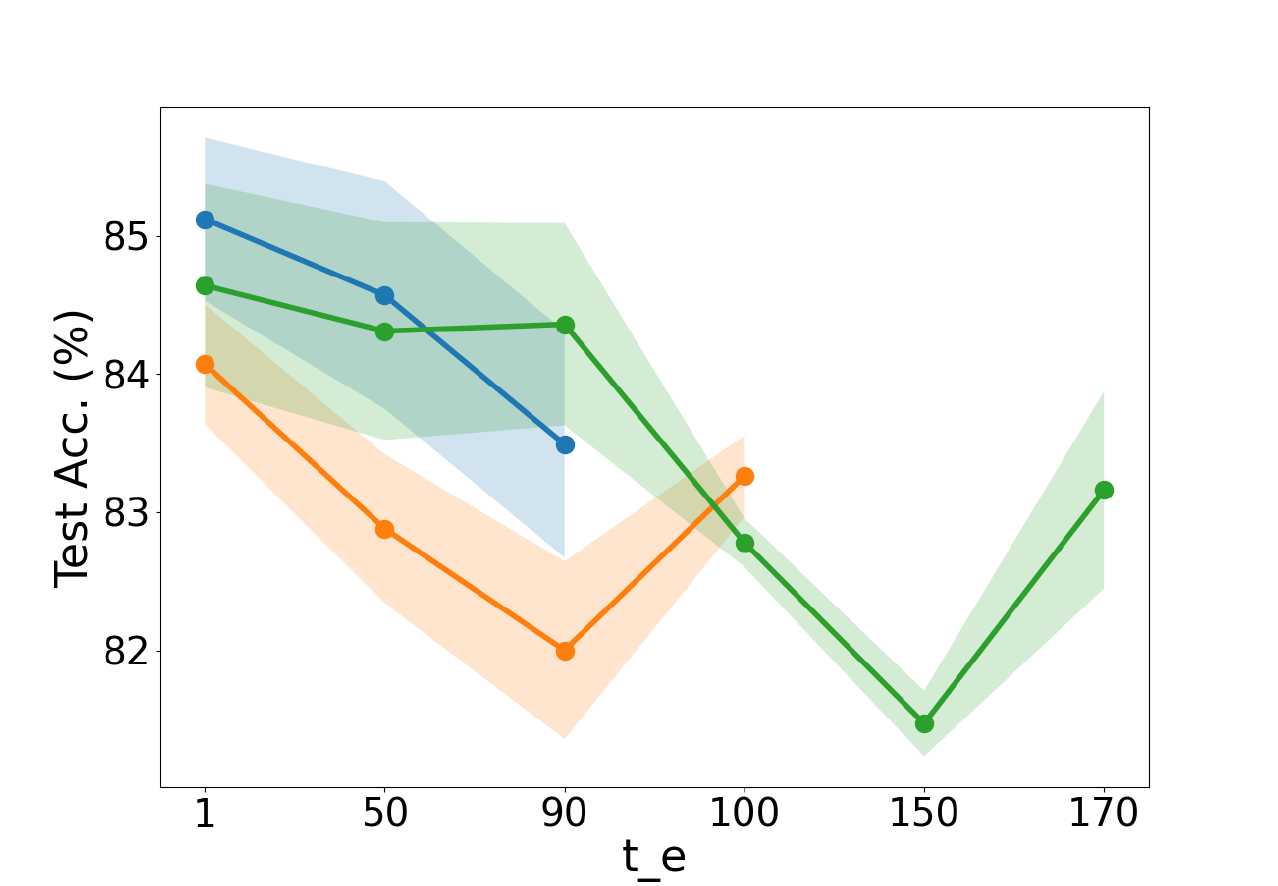} \\
        \vspace{1pt}
        \includegraphics[width=0.243\linewidth]{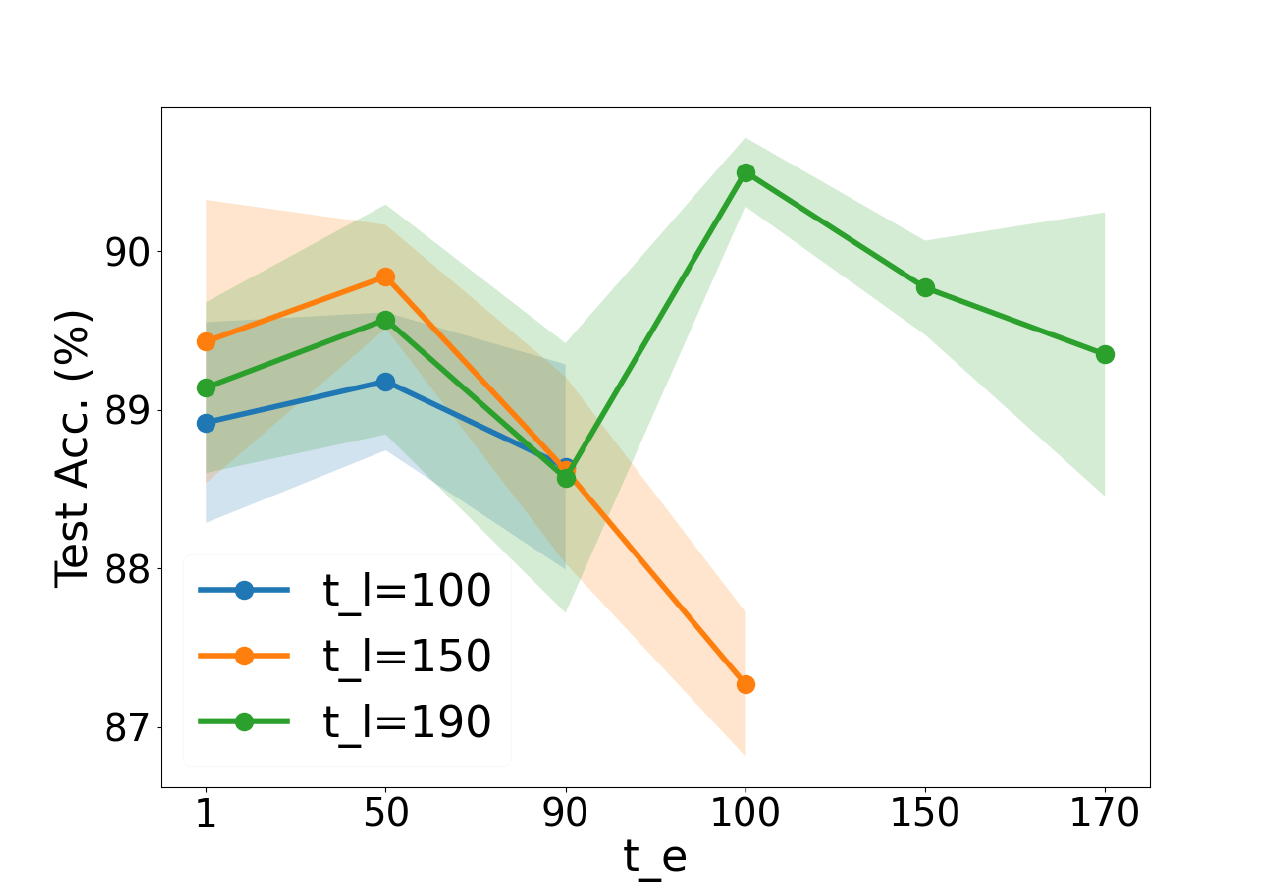}
        \includegraphics[width=0.243\linewidth]{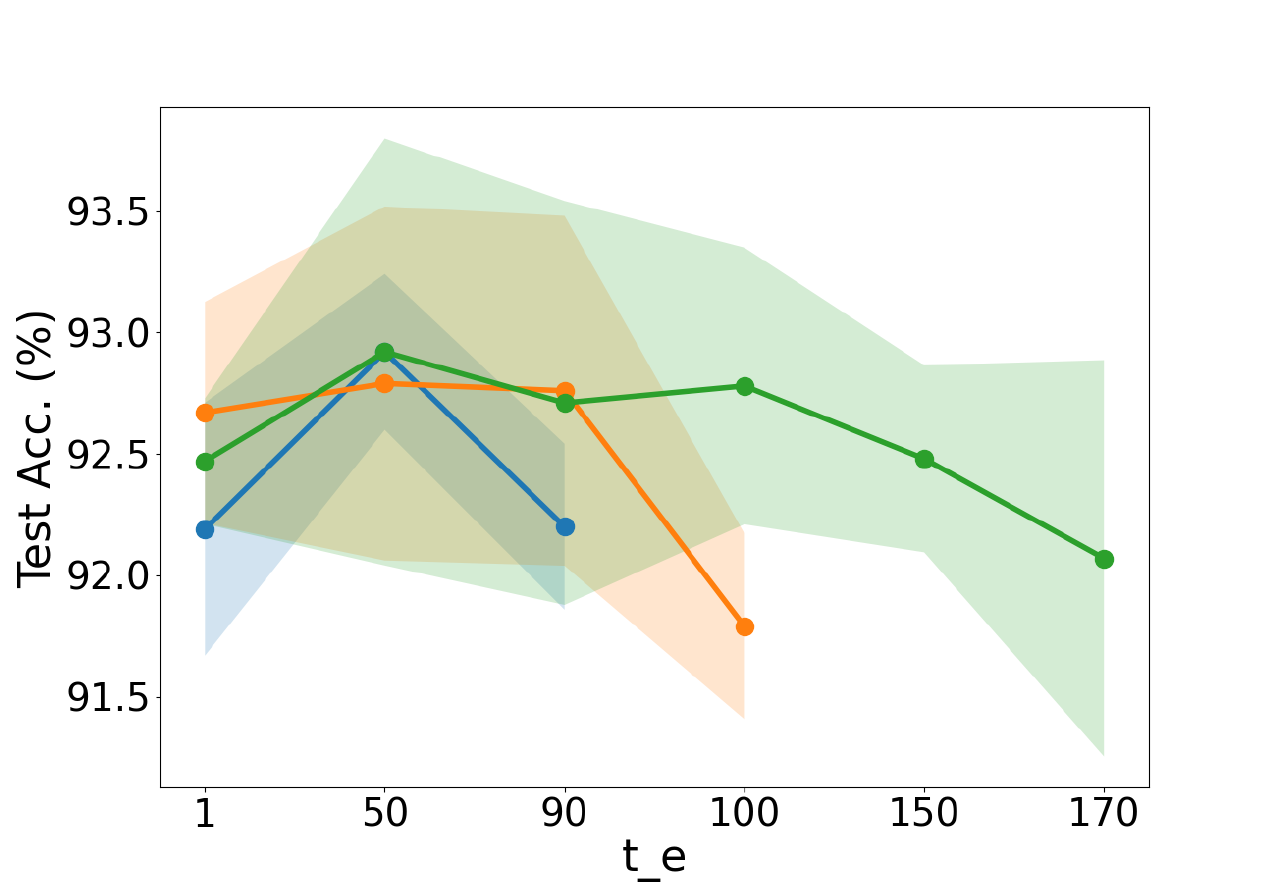} 
        \includegraphics[width=0.243\linewidth]{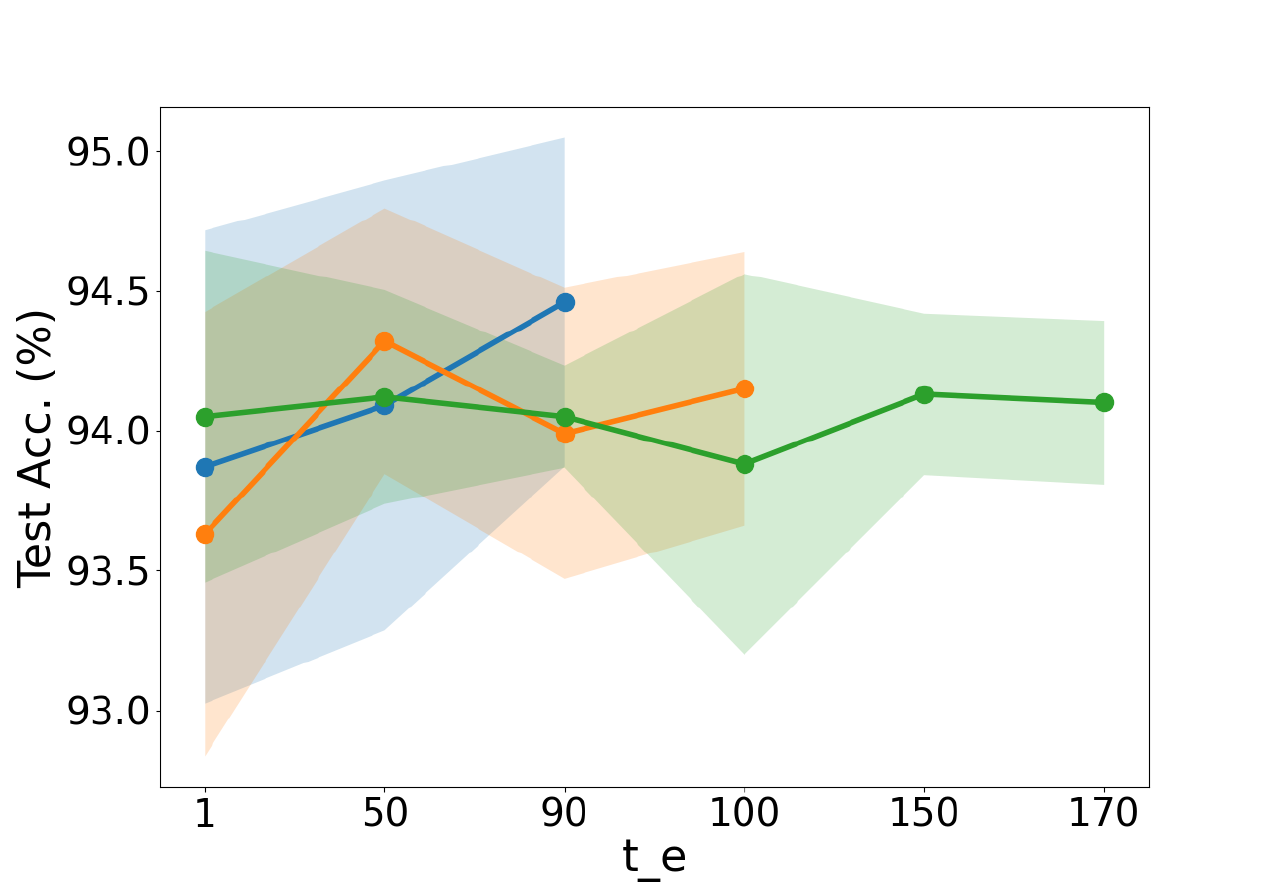}
        \includegraphics[width=0.243\linewidth]{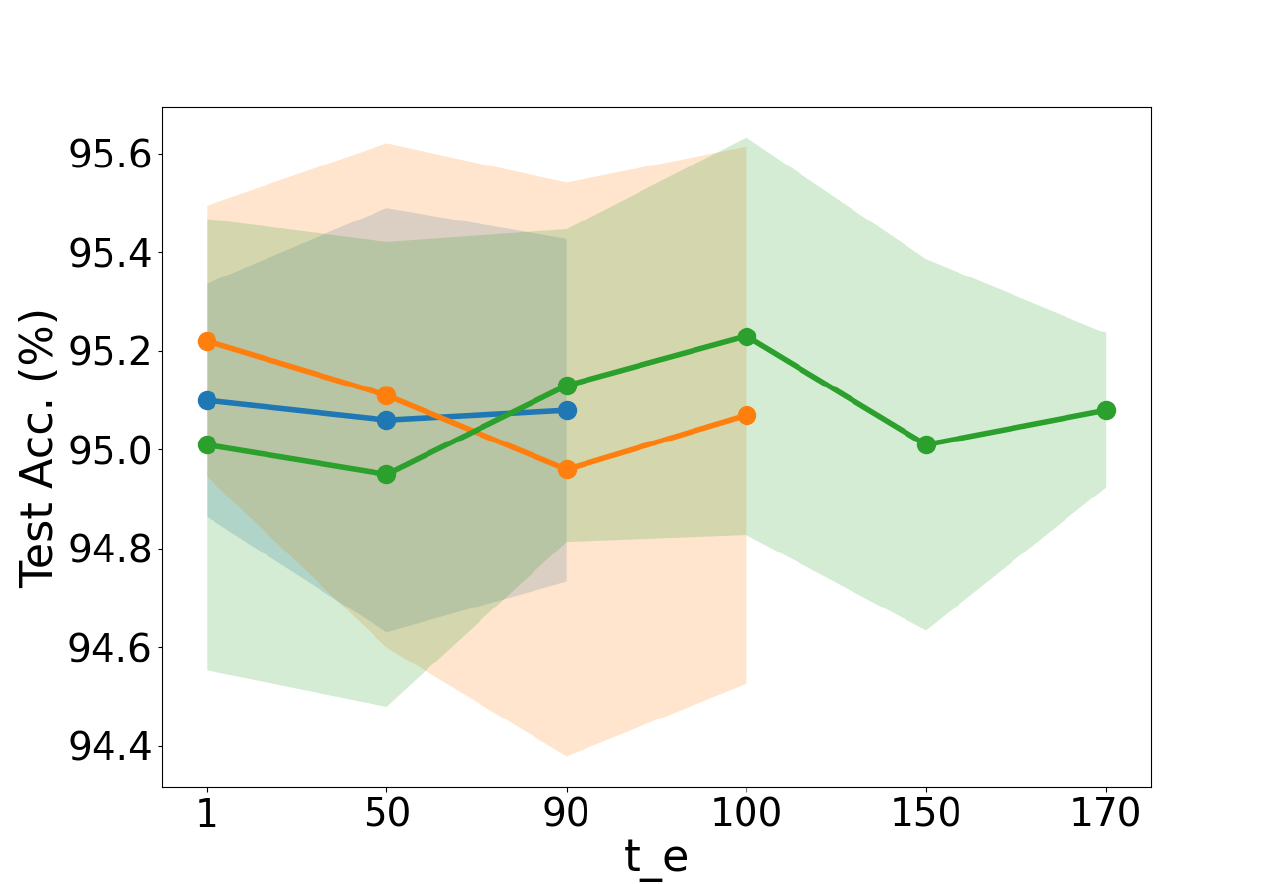}
        \caption{Window combinations on CIFAR-10. Different colors indicate the start epoch of different late windows $t_l$, and x-axis represents the start epoch of the early window $t_e$. From left to right and top to bottom, the corresponding selection rates are 0.02, 0.05, 0.1, 0.2, 0.3, 0.5, 0.7, and 0.9.}
        \label{fig:window_cifar}
    \end{figure*}

%% file: supplementary/tab_cifar.tex
%Table 1. Comparison of different methods.

% start: hf's tables
% problem: resizebox, rowcolor

\begin{table*}[htbp]
    \centering
    \caption{Performances on natural image dataset CIFAR-10 with ResNet-18. The model trained with the full dataset achieves 93.06\% accuracy. %The first and second best results in each column are marked in red and blue, respectively. 
    %All training is repeated 3 times with different random seeds to calculate mean accuracy with standard deviation. 
    }
    % \vspace{-5pt} %vertical space before the table 
    \resizebox{1\textwidth}{!}{ % ! scale in height automatically.
        \begin{large}
        \begin{tabular}{p{2.2cm}<{\centering}|c|c|c|c|c}
        \toprule[1.2pt]
        $\alpha$ & 2\% & 5\% & 10\% & 20\% & 30\%\\
        \midrule[0.6pt]
        \midrule[0.6pt]
        Random      & 41.64 $\pm$ 0.92\%   & \color{myblue}{58.62 $\pm$ 0.29\%} & 71.64 $\pm$ 0.47\%   & 84.57 $\pm$ 0.33\%  & 89.79 $\pm$ 0.32\% 
        \\
        Forgetting     & 36.20 $\pm$ 0.24\%   & 41.67 $\pm$ 0.58\% & 52.29 $\pm$ 0.33\%  &  76.00 $\pm$ 1.45\%  & \color{myblue}{90.27 $\pm$ 0.88\%}  
        \\
        Entropy     & 32.08 $\pm$ 0.42\%   & 47.70 $\pm$ 0.39\%& 60.52 $\pm$ 0.15\%   &    75.69 $\pm$ 0.90\%  & 86.46 $\pm$ 0.57\%  
        \\
        EL2N      & 10.54 $\pm$ 0.49\%   & 15.94 $\pm$ 0.61\% & 23.45 $\pm$ 0.86\%   &    42.62 $\pm$ 0.68\%  & 81.69 $\pm$ 1.27\%  
        \\
        AUM       & 14.66 $\pm$ 0.52\%   & 18.54 $\pm$ 0.59\% & 25.35 $\pm$ 0.22\%   &    49.51 $\pm$ 1.10\%  & 73.92 $\pm$ 0.72\%  
        \\
        CCS      & \color{myblue}{43.95 $\pm$ 1.66\%}   & 51.45 $\pm$ 2.18\% & \color{myblue}{71.78 $\pm$ 1.98\%}   &    \color{myred}{85.53 $\pm$ 0.97\%}  & 89.70 $\pm$ 0.65\%  
        \\
        % \rowcolor{blue!20}
        \midrule[0.6pt]
        \rowcolor[HTML]{EFEFEF} 
        \textbf{\method(Ours)}   & \textbf{\color{myred}{46.27 $\pm$ 0.37\%}}   & \textbf{\color{myred}{61.75 $\pm$ 0.57\%}}   & \textbf{\color{myred}{73.73 $\pm$ 0.42\%}}  & \textbf{\color{myblue}{85.12 $\pm$ 0.68\%}}   & \textbf{\color{myred}{90.50 $\pm$ 0.49\%}} \\
        \bottomrule[1.2pt]
        \end{tabular}
        \end{large}}
    \vspace{10pt}
    \label{tab:cifar10}
\end{table*}

\begin{table*}[htbp]
    \centering
    \caption{Performances on natural image dataset CIFAR-100 with ResNet-18. The model trained with the full dataset achieves 78.46\% accuracy. % The first and second best results in each column are marked in red and blue, respectively. 
    }
    % \vspace{-5pt} %vertical space before the table 
    \resizebox{1\textwidth}{!}{ % ! scale in height automatically.
        \begin{large}
        \begin{tabular}{p{2.2cm}<{\centering}|c|c|c|c|c}
        \toprule[1.2pt]
        $\alpha$ & 2\% & 5\% & 10\% & 20\% & 30\%\\
        \midrule[0.6pt]
        \midrule[0.6pt]
        Random      & \color{myred}{13.35 $\pm$ 0.39\%}   & 20.53 $\pm$ 0.93\% & \color{myblue}{37.10 $\pm$ 1.01\%}   & 53.68 $\pm$ 1.33\%  & 62.74 $\pm$ 0.15\% 
        \\
        Forgetting    & 6.86 $\pm$ 0.08\%   & 10.14 $\pm$ 0.32\% & 16.87 $\pm$ 0.12\%  &  26.18 $\pm$ 0.61\%  & 38.25 $\pm$ 0.69\%  
        \\
        Entropy     & 8.92 $\pm$ 0.40\%   & 14.64 $\pm$ 0.50\%  & 25.01 $\pm$ 0.46\%   &    40.33 $\pm$ 0.24\%  & 48.95 $\pm$ 0.46\%  
        \\
        EL2N      & 3.63 $\pm$ 0.02\%   & 5.16 $\pm$ 0.22\% & 7.26 $\pm$ 0.22\%   &    14.65 $\pm$ 0.87\%  & 34.83 $\pm$ 0.50\%  
        \\
        AUM       & 3.92 $\pm$ 0.03\%   & 5.25 $\pm$ 0.04\%  & 8.38 $\pm$ 0.29\%   &    16.64 $\pm$ 0.07\%  & 31.34 $\pm$ 0.49\%  
        \\
        CCS       & 13.50 $\pm$ 0.47\%   & \color{myblue}{23.84 $\pm$ 1.07\%}  & 36.39 $\pm$ 1.94\%   &   \color{myblue}{53.14 $\pm$ 1.34\%}  & \color{myred}{64.72 $\pm$ 0.21\%}
        \\
        % \rowcolor{blue!20}
        \midrule[0.6pt]
        \rowcolor[HTML]{EFEFEF} 
        \textbf{\method(Ours)}   & \textbf{\color{myblue}{13.28 $\pm$ 0.33\%}}   & \textbf{\color{myred}{24.38 $\pm$ 0.87\%}}   & \textbf{\color{myred}{39.60 $\pm$ 0.46\%}}  & \textbf{\color{myred}{55.86 $\pm$ 0.92\%}}   & \textbf{\color{myblue}{62.93 $\pm$ 0.37\%}} \\
        \bottomrule[1.2pt]
        \end{tabular}
        \end{large}}
    \vspace{10pt}
    \label{tab:cifar100}
\end{table*}

%% file: supplementary/cross.tex
\section{Generalization across Architecture}
    In this section, we investigate the generalization ability across architectures.
    Specifically, We train a ResNet-18 model with the entire dataset and use various scores to select coresets with different selection rates. Then we train three representative architectures including ResNet-50, MobileNet-v2 and LeNet models with these coresets. 
    The evaluation results in \cref{tab:cross} demonstrate that the coresets selected by the proposed \method outperform the compared SOTA baselines and have good transferability across architectures.
    \input{supplementary/tab_cross}

%% file: supplementary/tab_cross.tex
\begin{table*}[h]
\renewcommand{\arraystretch}{0.95}
	\caption{Cross-architecture generalization performance. We train ResNet-50, MobileNet-v2 and LeNet models with coresets of OrganSMNIST selected by scores calculated on training dynamics with ResNet-18. %The first and second best results in each column are marked in red and blue, respectively.
 }
	\label{tab:cross}
	\begin{center}
        \vspace{-5pt}
        \resizebox{1\textwidth}{!}{
		\begin{Large}
			\setlength{\tabcolsep}{1.5mm}
			\begin{tabular}{c|cccc|cccc|cccc}
				\hline
                    \toprule[0.8pt]
                    % \rowcolor[HTML]{EFEFEF}
                    & \multicolumn{4}{c}{\textbf{ResNet-50}} &\multicolumn{4}{c}{\textbf{MobileNet-v2}} &\multicolumn{4}{c}{\textbf{LeNet}}\\ 
                    % \cmidrule(r){2-7} \cmidrule(r){7-11} 
                    $\alpha$ &{5\%} &{10\%} &{20\%} &{30\%} &{5\%} &{10\%} &{20\%} &{30\%} &{5\%} &{10\%} &{20\%} &{30\%}\\
                    \midrule[0.6pt]
                    \midrule[0.6pt]
                    % Full dataset   &\multicolumn{4}{c}{{98.39}} &\multicolumn{4}{c}{{98.39}}  &\multicolumn{4}{c}{{91.76}}\\
                    % \midrule[0.6pt]
                    
                    Random     & 23.73    & \color{myblue}76.81   & \color{myblue}82.13   & \color{myblue}83.79   & \color{myblue}48.10    & 77.64    & \color{myblue}83.45    & \color{myblue}86.91    & 49.95   & 62.99  & \color{myblue}67.09  & \color{myblue}79.35  \\
                    Forgetting   & 4.64     & 35.79   & 59.42   & 68.21   & 4.10     & 28.66    & 53.47    & 76.61    & 4.59    & 24.76  & 33.59  & 63.62  \\
                    Entropy   & 26.71    & 48.29   & 76.90    & 81.05   & 24.80     & 54.15    & 73.00    & 84.91    & 21.58   & 47.56  & 65.92  & 72.56  \\
                    EL2N      & 15.33    & 56.64   & 67.24   & 78.71   & 20.56    & 52.34    & 70.56    & 79.98    & 14.60   & 46.92  & 61.08  & 68.26  \\
                    AUM     & 4.10     & 22.36   & 37.16   & 53.81   & 4.00     & 21.44    & 37.55    & 58.01    & 4.30    & 13.38  & 32.08  & 42.82  \\
                    CCS       & \color{myblue}43.90    & 71.88   & 78.27   & 82.52   & 45.65    & \color{myblue}77.98    & 81.40    & 84.91    & \color{myblue}52.15   & \color{myblue}69.48  & 71.44  & 77.10  \\
                    \midrule[0.6pt]
                    \rowcolor[HTML]{EFEFEF}
                    \textbf{\method (Ours)} & \textbf{\color{myred}52.83}   & \textbf{\color{myred}77.00}   & \textbf{\color{myred}82.96}   & \textbf{\color{myred}85.84}   & \textbf{\color{myred}51.17}   & \textbf{\color{myred}79.79}   & \textbf{\color{myred}84.38}   & \textbf{\color{myred}88.43}   & \textbf{\color{myred}59.67}   & \textbf{\color{myred}70.90}   & \textbf{\color{myred}77.00}   & \textbf{\color{myred}80.32} \\
                    
                    \bottomrule[0.8pt]
				\hline
			\end{tabular}
		\end{Large}
  }
  \end{center}
\end{table*}

%% file: supplementary/entire.tex
\section{Comparison between Dual-Window and Entire-Process}
\begin{figure}[h]
        \centering
        \vspace{-5pt}
        \subfloat[]{
                \label{fig:entire_S}
                \includegraphics[width=0.49\linewidth]{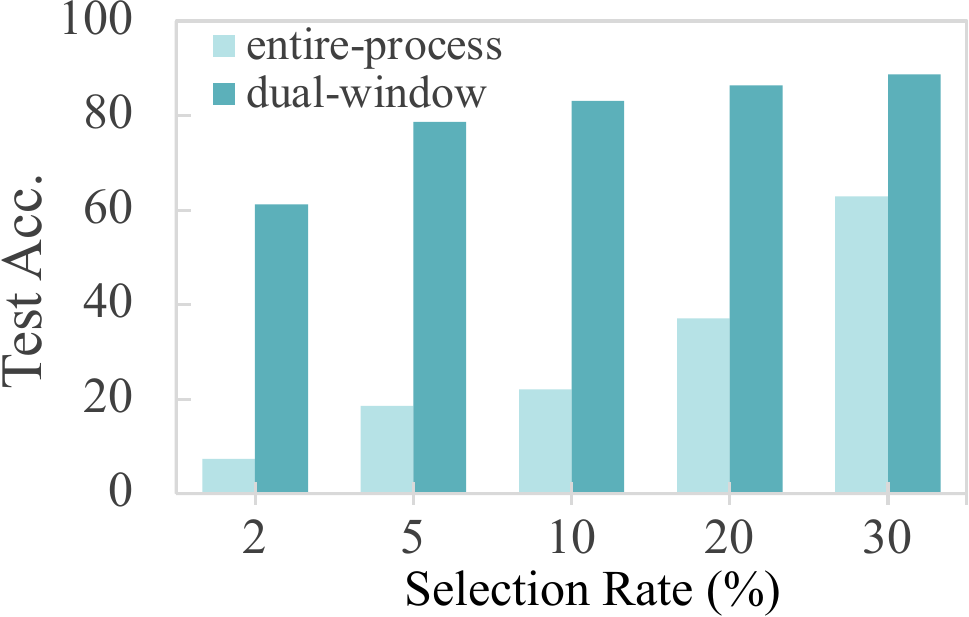}}
        \subfloat[]{
                \label{fig:entire_A}
                \includegraphics[width=0.49\linewidth]{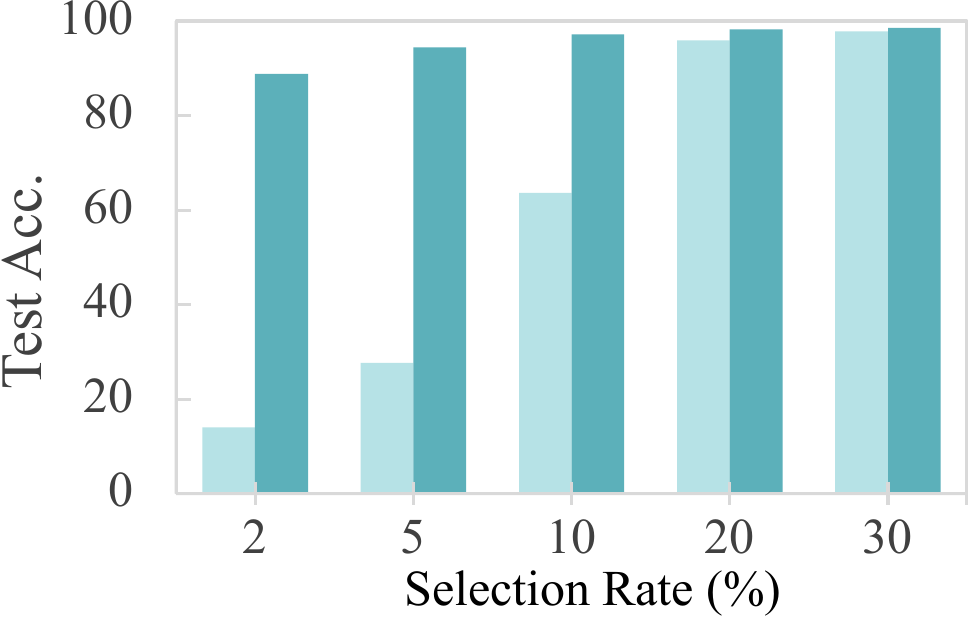}}
        \setlength{\abovecaptionskip}{8pt}
        % \end{minipage}
        % \captionsetup{labelformat=empty}
        \captionsetup{font={small,bf,stretch=1}, justification=raggedright}
        \captionof{figure}{Comparison between dual-window and entire-process.}
        \label{fig:entire}
    \end{figure}

    We compare the performance of calculating the dual-window variance with the entire-process variance (OrganSMNIST \cref{fig:entire_S}, OrganAMNIST \cref{fig:entire_A}).
    In most cases, the dual-window approach provides better performance than the entire-process method, because it effectively captures epochs that really matters. 
    For instance, some samples may show significant fluctuations at specific stages while stabilizing at the rest of the time. 
    The entire-process scheme diminishes the effect of these specific stages.

%% file: supplementary/parameter.tex
\section{Parameter Settings}
    \balance
    We train the ResNet-18 over 200 epochs with a batch size of 256. For networks update, SGD optimizer with momentum of 0.9 and weight decay of 0.0005 is used. The learning rate is initialized as 0.1 and decays with the cosine annealing scheduler.
    
    \noindent Besides, this section details the optimal window combinations identified for each dataset and selection rate assessed in our study.
    We represent each window combination of early and late stages, ($t_e$, $t_E$)+($t_l$, $t_L$), more concisely as ($t_e$, $t_l$), since we set the window size to $K=10$ throughout our experiments.
    For each dataset, We list the optimal ($t_e$, $t_l$) of every selection rate $\alpha$ as follows in the format of ($t_e$, $t_l, \alpha$). 

    \vspace{5pt}
    \setlist[itemize]{leftmargin=3mm}
    \begin{itemize}
    \setlength{\itemsep}{-1pt}
    \setlength{\parsep}{-1pt}
    \setlength{\parskip}{-1pt}
        \item For OrganAMNIST, the optimal settings are (1, 190, 2\%), (1, 190, 5\%), (100, 190, 10\%), (1, 150, 20\%), (90, 150, 30\%), (90, 150, 50\%), (1, 190, 70\%), (90, 190, 90\%). 
    \end{itemize}
    
    \begin{itemize}
        \item For OrganSMNIST, the optimal settings are (90, 100, 2\%), (150, 190, 5\%), (100, 190, 10\%), (100, 190, 20\%), (170, 190, 30\%), (150, 190, 50\%), (90, 150, 70\%), (1, 100, 90\%). 
    \end{itemize}
    
    \begin{itemize}
        \item For CIFAR-10, the optimal settings are (1, 100, 2\%), (1, 150, 5\%), (1, 190, 10\%), (1, 100, 20\%), (100, 190, 30\%).
    \end{itemize}
    
    \begin{itemize}
        \item For CIFAR-100, the optimal settings are (170, 190, 2\%), (100, 190, 5\%), (170, 190, 10\%), (100, 190, 20\%), (90, 190, 30\%). 
    \end{itemize}

    % \captionsetup[subfig]{labelformat=empty}
    % \setlength{\abovecaptionskip}{5pt}
    % \begin{figure*}[ht]
    %     \centering
    %     %\includegraphics[width=3in]{fig5}
    %     \includegraphics[width=0.243\linewidth]{supplementary/figs/demo-2.pdf} 
    %     \includegraphics[width=0.243\linewidth]{supplementary/figs/demo-5.pdf} 
    %     \includegraphics[width=0.243\linewidth]{supplementary/figs/demo-10.pdf} %\hspace{-1mm}
    %     \includegraphics[width=0.243\linewidth]{supplementary/figs/demo-20.pdf} \\
    %     \vspace{1pt}
    %     \includegraphics[width=0.243\linewidth]{supplementary/figs/demo-30.pdf}
    %     \includegraphics[width=0.243\linewidth]{supplementary/figs/demo-50.pdf} 
    %     \includegraphics[width=0.243\linewidth]{supplementary/figs/demo-70.pdf}
    %     \includegraphics[width=0.243\linewidth]{supplementary/figs/demo-90.pdf}
    %     \caption{Window combinations on CIFAR-10. Different colors indicate the start epoch of different late windows $t_l$, and x-axis represents the start epoch of the early window $t_e$. From left to right and top to bottom, the corresponding selection rates are 0.02, 0.05, 0.1, 0.2, 0.3, 0.5, 0.7, and 0.9.}
    %     \label{fig:window_cifar}
    % \end{figure*}
    
    \noindent As reported in \cref{fig:window_cifar}, we analyze the influence of window combination on performance. For a smaller selection rate, we should select samples earlier in training, and as the selection rate increase, the data budgets also increase, therefore the optimal window combination gradually slides from early to later stage.

%% file: supplementary/time.tex
\section{Analysis of the Computational Overhead}
    We provide the computational cost in \cref{tab:time} by comparing \method to other baselines in terms of the time it takes to compute the score (on an NVIDIA RTX 3090 GPU).
    We chose EL2N as the baseline since it only requires the first 10 epochs to generate the score. 
    % \method incurs additional computational overhead compared to EL2N due to the dual-window and variance calculations.
    The computational overhead incurred by dual-window and variance calculations are tolerable compared to the cost of training a modern deep neural network on the full training set, and smaller than most compared methods.
    \input{supplementary/tab_time}

%% file: supplementary/tab_time.tex
% problem: resizebox, rowcolor
% \vspace{10pt}
\begin{table}[htbp]
    % \vspace{-2.0em}
    \vspace{-8pt}
    \centering
    \captionsetup{font={small,bf,stretch=1}, justification=raggedright}
    \caption{Computational costs on calculating scores.}
    \vspace{-10pt}
    \resizebox{1\linewidth}{!}{ 
    % \begin{large}
    % \begin{tabular}{p{2cm}<{\centering}|ccc|ccc}
    \begin{tabular}{c|cc|cc|cc}
    \toprule[1.2pt]
    
    & \multicolumn{2}{c}{\textbf{OrganAMNIST}} &\multicolumn{2}{c}{\textbf{OrganSMNIST}} &\multicolumn{2}{c}{\textbf{PathMNIST}} \\ 
    & Time (sec) & $\Delta$ EL2N  & Time (sec) & $\Delta$ EL2N  & Time (sec) & $\Delta$ EL2N\\
    \midrule[0.6pt]
    \midrule[0.6pt]
    % TissueMNIST
    % EL2N         & 5.03 & -     & 2.08 & -     & 24.15 & -      \\
    % Forgetting   & 7.87 & +2.84 & 3.32 & +1.24 & 38.63 & +14.48 \\
    % Entropy      & 5.46 & +0.43 & 3.96 & +1.88 & 24.49 & +0.34  \\
    % AUM          & 9.53 & +4.50 & 4.10 & +2.02 & 46.73 & +22.58 \\
    % CSS (on AUM) & 9.55 & +4.52 & 4.11 & +2.03 & 46.80 & +22.65 \\
    % EL2N         & 12.26 & -       & 6.26 & -      & 49.95 & -       \\
    % Forgetting   & 15.10  & +23.2\% & 7.50  & 19.8\% & 64.43 & +29.0\% \\
    % Entropy      & 12.69 & +3.5\%  & 8.14 & 30.0\% & 50.29 & +0.7\%  \\
    % AUM          & 16.76 & +36.7\% & 8.28 & 32.3\% & 72.53 & +45.2\% \\
    % CSS (on AUM) & 16.78 & +36.9\% & 8.29 & 32.4\% & 72.60  & +45.3\% \\
    
    % PathMNIST
    EL2N         & 12.26 & -       & 6.26 & -      & 13.59 & -       \\
    Forgetting   & 15.10  & +23.2\% & 7.50  & 19.8\% & 20.33 & +49.6\% \\
    Entropy      & 12.69 & +3.5\%  & 8.14 & 30.0\% & 16.31 & +20.0\%  \\
    AUM          & 16.76 & +36.7\% & 8.28 & 32.3\% & 24.25 & +78.4\% \\
    CSS (on AUM) & 16.78 & +36.9\% & 8.29 & 32.4\% & 24.28  & +78.7\% \\

    \rowcolor[HTML]{EFEFEF}
    % \method  & 5.81 & +0.78 & 2.34 & +0.26 & 46.26 & +22.11 \\
    \textbf{\method (Ours)}  & \textbf{13.04} & \textbf{+6.4\% } & \textbf{6.52 }& \textbf{4.2\%}  & \textbf{17.53} & \textbf{+29.0\%} \\
\bottomrule[1.2pt]

\end{tabular}
% \end{large}
}
% \vspace{-10pt}
% \vspace{-1.5em}
\label{tab:time}
\end{table}